\documentclass[10pt,twocolumn,letterpaper]{article}

\usepackage[pagenumbers]{cvpr}      %

\definecolor{cvprblue}{rgb}{0.21,0.49,0.74}
\usepackage[pagebackref,breaklinks,colorlinks,allcolors=cvprblue]{hyperref}

\usepackage{multirow}
\usepackage{pifont}
\usepackage{makecell}
\usepackage{float}
\usepackage{afterpage}
\usepackage{balance}

\newtoggle{comments}
\toggletrue{comments}
\iftoggle{comments}{%
    \newcommand{\tarasha}[1]{{\leavevmode\color{magenta}[Tarasha: #1]}}
    \newcommand{\deva}[1]{{\leavevmode\color{blue}[Deva: #1]}}
    \newcommand{\kaihua}[1]{{\leavevmode\color{blue}[Kaihua: #1]}}
}{%
  \newcommand{\tarasha}[1]{}
  \newcommand{\deva}[1]{}
  \newcommand{\kaihua}[1]{}
}

\def\paperID{2807} %
\def\confName{CVPR}
\def\confYear{2025}

\title{Using Diffusion Priors for Video Amodal Segmentation}

\author{Kaihua Chen \hspace{0.5cm} Deva Ramanan \hspace{0.5cm} Tarasha Khurana\\
Carnegie Mellon University}

\begin{document}

\maketitle

\begin{abstract}
Object permanence in humans is a fundamental cue that helps in understanding persistence of objects, even when they are fully occluded in the scene. Present day methods in object segmentation do not account for this {\em amodal} nature of the world, and only work for segmentation of visible or {\em modal} objects. Few amodal methods exist; single-image segmentation methods cannot handle high-levels of occlusions which are better inferred using temporal information, and multi-frame methods have focused solely on segmenting rigid objects. To this end, we propose to tackle video amodal segmentation by formulating it as a conditional generation task, capitalizing on the foundational knowledge in video generative models. Our method is simple; we repurpose these models to condition on a sequence of modal mask frames of an object along with contextual pseudo-depth maps, to learn which object boundary may be occluded and therefore, extended to hallucinate the complete extent of an object. This is followed by a content completion stage which is able to inpaint the occluded regions of an object.
We benchmark our approach alongside a wide array of state-of-the-art methods on four datasets and show a dramatic improvement of upto 13\% for amodal segmentation in an object's occluded region. 
\end{abstract}    
\section{Introduction}
\label{sec:intro}

Gestalt psychology~\cite{koffka2013gestalt} suggests that human perception inherently organizes visual elements into cohesive wholes. When an object is occluded, humans can often infer the complete outline of the object -- an ability that is developed in humans in their early years ~\cite{kavsek2004influence, otsuka2006development}. Additionally, object permanence \cite{baillargeon1991object} suggests that with some temporal context, humans can perceive objects to \textit{persist} even when they undergo complete occlusions. Replicating these phenomena of gestalt psychology and object permanence in object segmentation has traditionally been ignored, as the community has focused largely on segmenting the visible or \textit{modal} regions of objects (as exemplified by models like SAM~\cite{kirillov2023sam, ravi2024sam2}). Recent focus has shifted to include amodal segmentation~\cite{li2016amodal, zhu2017semantic}, which involves segmenting an object's full shape, including both visible and occluded parts. This task has broad real-world applications, including safe navigation in robotic manipulation and autonomous driving~\cite{qi2019kins, reddy2022walt}, understanding occluder-occludee relationships in complex scenes~\cite{zhan2020pcnet}, and enhancing advanced image and video editing tools~\cite{ozguroglu2024pix2gestalt}.

{\bf Why is this hard?} In a monocular setup, amodal perception is an ill-posed problem because there are multiple plausible explanations for how an object boundary should be extended in an occluded region. Recent innovations for amodal segmentation~\cite{ozguroglu2024pix2gestalt, zhan2024sdamodal} and inpainting~\cite{lugmayr2022repaint, podell2023sdxl} use diffusion frameworks for learning this multi-modal distribution, but they are not able to handle scenarios where an object could be fully-occluded. This issue is exacerbated by the lack of real-world datasets that have groundtruths for both amodal masks of objects, and their RGB content.

{\bf Status quo.} Despite this, current image-based amodal segmentation algorithms~\cite{zhan2020pcnet, ke2021bcnet, tran2022aisformer, gao2023c2fseg, ozguroglu2024pix2gestalt, zhan2024sdamodal, tran2024amodal} have shown impressive performance. However, these approaches are set in the single-frame setting, where they struggle with cases where objects are heavily or completely occluded. A potential solution is to approach amodal segmentation in a multi-frame setting~\cite{khurana2021detecting} so as to infer complete occlusions with temporal context. 
However, existing \textit{video} amodal segmentation algorithms~\cite{yao2022savos, fan2023eoras} are typically limited to rigid objects, and are dependent on additional inputs (like camera poses or optical flow) which hinders their scalability and therefore, generalization to unseen data.

{\bf Key insight.} To address these challenges, we propose repurposing a video diffusion model, Stable Video Diffusion (SVD)~\cite{svd}, to achieve highly accurate and generalizable video amodal segmentation. One key insight is that foundational diffusion models trained to generate pixels also bake-in strong priors on object shape. Such priors have been expoited by conditional image generation~\cite{ramesh2021dalle, saharia2022imagen, zhang2023controlnet} methods that condition on semantic maps and object boundaries. We similarly exploit these priors for our task. But crucially, our multi-frame video setup allows us to propagate object shape and content across time; %
e.g., one can infer the shape of a fully occluded by object by looking at {\em other} frames where it is visible (Fig.~\ref{fig:0}). %

Our proposed model achieves state-of-the-art performance across four synthetic and real-world video datasets, compared to a wide-variety of single-frame and multi-frame amodal segmentation baselines. We train on only synthetic data, but demonstrate strong zero-shot generalization to real-world data.
Thanks to the multi-modal generation capability of diffusion models, our approach can provide multiple plausible interpretations for the completion of occluded objects. We show that the outputs of our approach can be used for downstream applications like 4D reconstruction, scene manipulation, and pseudo-groundtruth generation.

\section{Related Work}
\label{sec:related}

\textbf{Image amodal segmentation.} Most previous amodal segmentation research has concentrated on image-based approaches. Some methods~\cite{li2016amodal, follmann2019orcnn, qi2019kins, 
xiao2021vrsp, ke2021bcnet, tran2022aisformer, tran2024amodal} adopt a similar strategy to modal segmentation, where models are trained to take RGB images as input and directly output amodal masks for all objects in the scene. Another line of methods~\cite{ehsani2018segan, ling2020amodalvae, zhan2020pcnet, ozguroglu2024pix2gestalt, zhan2024sdamodal, xu2024amodal} leverages existing modal masks, generated by modal segmentation models, to predict amodal masks based on these and additional inputs like image frames. Besides inferring the complete object shape, some approaches also \textit{hallucinate} the RGB content in the occluded regions. Generic inpainting methods~\cite{lugmayr2022repaint, podell2023sdxl} often fail at this task, as they rely on surrounding context, which often includes occluders. In contrast, content completion methods explicitly condition on the modal content, either by directly generating the amodal content based on modal information~\cite{ozguroglu2024pix2gestalt, xu2024amodal} or by inpainting within the predicted amodal segmentation area~\cite{ling2020amodalvae, zhan2020pcnet}. Due to the availability of high-quality real-world amodal image datasets~\cite{ martin2001database, zhu2017semantic,   qi2019kins}, image amodal segmentation and content completion methods have shown strong performance by learning robust shape priors. However, these methods frequently struggle with cases of significant occlusion and fail entirely for fully occluded objects because the amodal cues cannot be inferred in a single-frame setting.

\textbf{Video amodal segmentation.} 
Recently, \textit{video} amodal segmentation methods have emerged~\cite{gao2023c2fseg, yao2022savos, fan2023eoras}. These approaches integrate information from preceding and succeeding frames in a video sequence, enabling temporally consistent predictions. However, the training and evaluation of most of these algorithms are limited to synthetic datasets with rigid objects of similar scale~\cite{greff2022kubric, tangemann2021fishbowl, geiger2013kitti, qi2019kins}. Although these algorithms outperform image-based amodal segmentation methods within synthetic datasets, their practical applications remain limited. In contrast, we utilize both synthetic~\cite{hu2019sailvos, greff2022kubric} and real-world datasets~\cite{dave2020tao, athar2023burst, hsieh2023taoamodal}, which include deformable objects with diverse motions and scales, often mixed with complex camera movements. Moreover, to our knowledge, this work is the first to explore video-level amodal \textit{content completion}.

\begin{figure*}[t]
  \centering
  \includegraphics[width=0.9\linewidth]{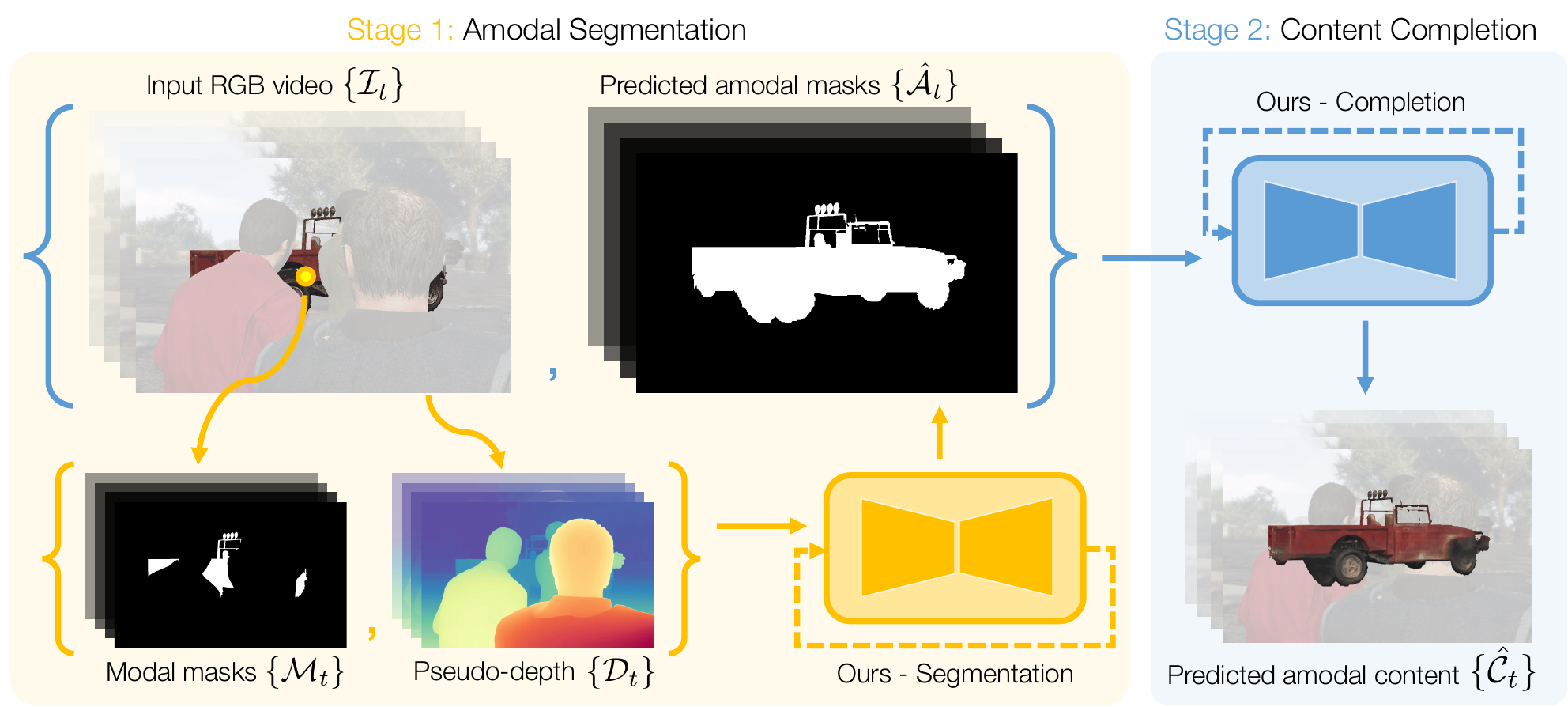}
  
  \captionsetup{justification=justified, singlelinecheck=false}
  \caption{\textbf{Model pipeline} for amodal segmentation and content completion. The first stage of our pipeline generates amodal masks $\{\hat{\mathcal{A}}_t\}$ for an object, given its modal masks $\{\mathcal{M}_t\}$ and pseudo-depth of the scene $\{\mathcal{D}_t\}$ (which is obtained by running a monocular depth estimator on RGB video sequence $\{\mathcal{I}_t\}$ ). The predicted amodal masks from the first stage are then sent as input to the second stage, along with the modal RGB content of the occluded object in consideration. The second stage then inpaints the occluded region and outputs the amodal RGB content $\{
\hat{\mathcal{C}}_t\}$ for the occluded object. Both stages employ a conditional latent diffusion framework with a 3D UNet backbone \cite{svd}. Conditionings are encoded via a VAE encoder into latent space, concatenated, and processed by a 3D UNet with interleaved spatial and temporal blocks. CLIP embeddings of $\{\mathcal{M}_t\}$ and the modal RGB content %
  provide cross-attention cues for the first and second stage respectively. Finally, the VAE decoder translates outputs back to pixel space.}
  \label{fig:1}
\end{figure*}

\textbf{Real-world priors from diffusion models.}
Diffusion models have achieved significant success in generative tasks within computer vision. Initially developed for unconditional image generation~\cite{ddpm}, the scope of diffusion models has expanded in multiple directions. These advancements include, but are not limited to, implementing conditional techniques for tasks like style transfer~\cite{zhang2023controlnet, brooks2023instructpix2pix} and text-to-image synthesis~\cite{ramesh2021dalle}, transitioning from pixel-space noise to latent-space noise~\cite{ldm, vldm}, developing various training and sampling strategies~\cite{ddpm, song2020ddim, edm}, and extending their application from realistic image generation to video generation~\cite{snapv,svd,ho2022imagenv}. In addition, the successful adaptation of diffusion models for multiple downstream tasks, including depth estimation~\cite{ke2024marigold}, multi-view synthesis~\cite{sargent2023zeronvs, liu2023zero}, and scene reconstruction~\cite{liu2024reconx}, underscores their ability to capture object shape priors and understand potential 3D information~\cite{zhan2023general}. While recent image amodal segmentation methods have demonstrated initial success in incorporating diffusion models~\cite{ozguroglu2024pix2gestalt, zhan2024sdamodal, xu2024amodal, tran2024amodal}, our approach advances this progress by applying video diffusion techniques to the domain of video amodal segmentation.

\section{Method}

Consider a video sequence $\{\mathcal{I}_1, \mathcal{I}_2, \dots,  \mathcal{I}_T \}$ with modal (or visible) segmentation masks  $\{\mathcal{M}_1, \mathcal{M}_2, \dots, \mathcal{M}_T\}$ for a target object. %
Such masks can be readily obtained by conventional modal segmentors, such as Segment Anything v2~\cite{ravi2024sam2}. %
We first describe a (diffusion-based) model  to generating amodal masks $\{\mathcal{A}_1, \mathcal{A}_2, \dots, \mathcal{A}_T\}$ that capture the full extent of the target object, including occluded portions. %
We then train a second stage (diffusion-based) model that uses the input video and amodal masks to fill in (or inpaint) the RGB content of the occluded areas $\{\mathcal{C}_1, \mathcal{C}_2, \cdots, \mathcal{C}_T\}$.

\subsection{Preliminary: diffusion framework} 

We make use of an open-source video latent diffusion model ~\cite{ldm, vldm} (Stable Video Diffusion (SVD)~\cite{svd}) and use the EDM framework~\cite{edm} for both training and inference. Compared to pixel-space diffusion, latent diffusion models reduce computational and memory demands by encoding frames into compact latent representations while preserving both perceptual and region-based alignment. The EDM framework further accelerates training convergence and reduces the required number of denoising steps during inference without compromising generation quality.

Our diffusion model takes as input the latent representation $\mathbf{z_0}$ , additional conditioning $\mathbf{c}$, a noise scale $\sigma$ following $\log \sigma \sim \mathcal{N}(P_{mean}, P_{std}) $, and Gaussian noise $\epsilon \sim \mathcal{N}(0,\sigma^2 I)$. The training objective is defined as:

\begin{equation}
\min_{\theta} \mathbb{E}_{\sigma, \mathbf{z_0}, \mathbf{c}, \epsilon} \left[ \lambda || D_{\theta}(\mathbf{z_0}+\epsilon; \sigma, \mathbf{c}) - \mathbf{z_0} ||^2_2 \right] 
\label{eq1}
\end{equation}

Here, $\lambda$ is a scalar related to $\sigma$, and $D_{\theta} = c_1 (\mathbf{z_0}+\epsilon) + c_2 F_{\theta}(\mathbf{z_0}+\epsilon; \sigma, \mathbf{c})$ represents the predicted latent representation, which combines the noisy latent input with the v-prediction~\cite{vpred} output of the diffusion backbone  $F_{\theta}$, using additional scalars $c_1$ and $c_2$ that also depends on $\sigma$.

\subsection{Modal masks in, amodal masks out}
To train a high-quality amodal segmentor with limited data, one strategy is to leverage the shape and content priors of video foundation models pretrained on large-scale datasets. For this, we lean on the foundational knowledge in SVD, learnt by pretraining on the extensive LDM-F dataset~\cite{svd} with 152 million examples. However, as the vanilla SVD was designed for image-to-video tasks, we adapt its structure and conditioning to suit our modal-to-amodal sequence generation task. We describe this below.

First, we replace the input conditioning $\mathbf{c}$,  originally an RGB image, with binary modal masks of shape $\mathcal{R}^{T \times 1 \times H \times W}$. By default, the variational autoencoder (VAE)~\cite{kingma2013vae} in SVD requires a 3-channel input. To address this mismatch in the number of channels, we replicate the binary mask three times, following the approach for single-channel VAE inputs in a recent work~\cite{ke2024marigold}. After encoding each (replicated) mask seperately, we obtain a latent tensor of shape $\mathcal{R}^{T \times C_1 \times \frac{H}{F} \times \frac{W}{F}}$. This latent representation, concatenated with a noise image of the same shape, forms the input to our backbone which is a spatio-temporal 3D U-Net~\cite{ronneberger2015unet, vldm}. The final shape of this input becomes $\mathcal{R}^{T \times 2C_1 \times H \times W}$. In contrast to the vanilla SVD, where the latent space of a single image is duplicated $T$ times to align with the 3D U-Net's input requirements, our 3D U-Net gets as input $T$ \textit{unique} frames of the modal mask sequence being used as conditioning.

Additionally, we use CLIP embeddings~\cite{radford2021clip} for the modal masks, and inject them into the transformer layers for cross-attention. This provides temporal information about the visibility of objects in surrounding frames. After the 3D U-Net, the VAE decoder converts the latent amodal mask predictions back into the pixel space.

\subsection{Conditioning on pseudo-depth}
Till now, we described how SVD is modified to enable predicting amodal masks from modal masks. 
We find that one can add more contextual cues about the object and scene in consideration through different data modalities. A natural choice for conditioning is RGB frames, as used in previous work~\cite{zhan2020pcnet, ozguroglu2024pix2gestalt}. However, since occlusions of the target object are typically caused by objects closer to the camera, we empirically find that pseudo-depth maps provide more implicit clues about potential occluders than RGB frames, making them a more effective indicator for determining regions to complete. We demonstrate the advantages of pseudo-depth over RGB conditioning in our ablation study. To integrate this, we utilize the Depth Anything V2 monocular depth estimator~\cite{yang2024depthanythingv2} to convert RGB images into pseudo-depth maps, which are then incorporated into our video diffusion model as additional channels concatenated to the aforementioned input.

With the addition of pseudo-depth conditioning, the input latents for our 3D U-Net backbone have the shape $\mathcal{R}^{T \times 3C_1 \times \frac{H}{F} \times \frac{W}{F}}$, requiring a new first convolutional layer in the 3D U-Net to accommodate the increased channels. Rather than finetuning our model with both modal masks and pseudo-depth conditionings directly, we find that it is more efficient to do a two-stage finetuning, where we finetune our mask conditioned model first and then use it to initialize the finetuning of the mask-and-depth conditioned model. We call this approach \textit{two-stage finetuning}, allowing the model to adapt gradually to the new conditions.

Inspired by ControlNet~\cite{zhang2023controlnet}, we retain the parameters of the first channels $2C_1$ in the input layer from the previously trained model and initialize the newly added channels $C_1$ to zero. This \textit{zero convolution} approach ensures the model retains its initial predictive capability during the first few fine-tuning steps with the added pseudo-depth conditioning. We demonstrate the importance of these training strategies in the ablation study.

\subsection{Amodal content completion}

\begin{figure}[t]
  \centering
   \includegraphics[width=1.0\linewidth]{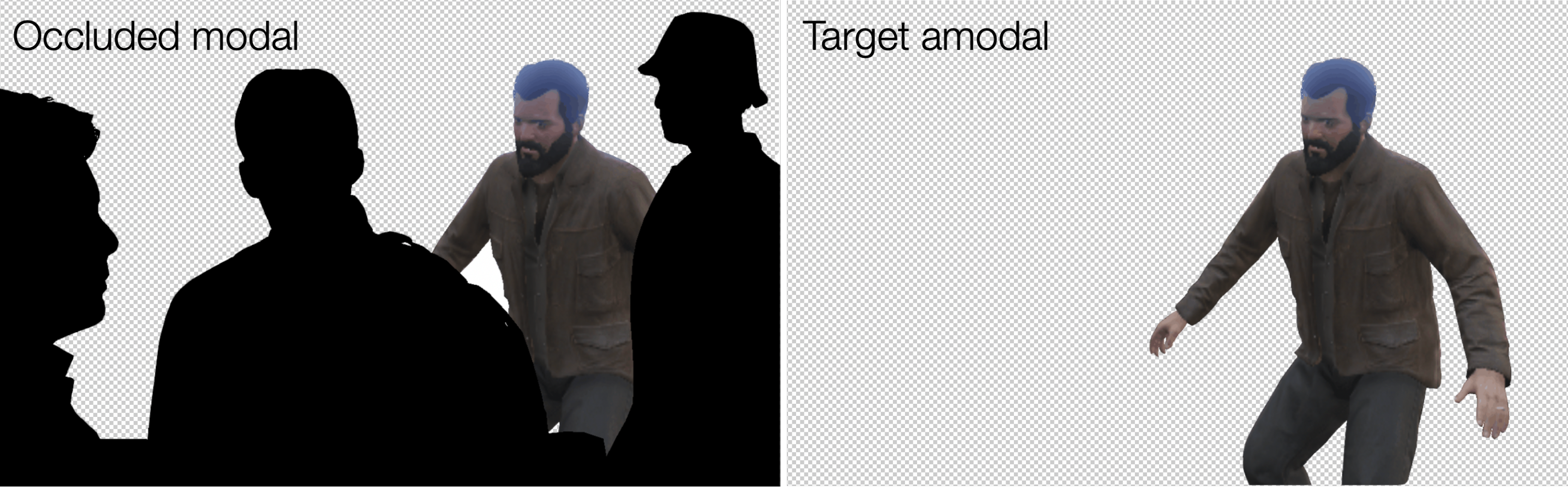}

   \caption{\textbf{Modal-amodal RGB training pair} for content completion. The left frame displays the partially occluded modal RGB content, generated by overlaying amodal masks (black regions) onto the amodal object to disrupt its visual integrity. The right frame shows the original, unoccluded amodal RGB object.}
   \label{fig:data_pair}
\end{figure}

Till now, we discussed the first stage of our pipeline which outputs amodal masks for occluded objects. However, the RGB content in the occluded region is unknown. To \textit{inpaint} these occluded areas, we use a second SVD model with the same architecture but with different conditionings; the first conditioning is the RGB content from an object's modal region, and the second conditioning is the predicted amodal mask from the first stage. We train this model to generate RGB content across the entire amodal region.

{\bf Synthetic data curation} A key challenge with this approach is the lack of ground-truth RGB content in occluded regions, even in synthetic datasets like SAIL-VOS~\cite{hu2019sailvos}. Inspired by self-supervised training-pair construction used extensively in image amodal tasks~\cite{zhan2020pcnet, ozguroglu2024pix2gestalt}, we extend this approach to video sequences. Figure~\ref{fig:data_pair} illustrates an example of a modal-amodal RGB content training pair. To construct such a pair, we first select an object from the dataset with near-complete visibility (above 95\%). We then sequentially overlay random amodal mask sequences onto this fully visible object until its visibility falls below a set threshold, thereby simulating occlusion. This effectively generates ground-truth RGB data for the occluded regions.

\section{Experiments}

\subsection{Setup}

\textbf{Implementation details.} 
For training, we load the official SVD-xt 1.1 pretrained checkpoint and use the AdamW optimizer with $\beta_1$=0.9, $\beta_2$=0.999. The learning rates for the two-stage fine-tuning are set to $3\cdot 10^{-5}$ and $3\cdot 10^{-6}$, for training without and with additional pseudo-depth conditioning, respectively. In the case of SAIL-VOS, due to computational limitations, we set the batch size to 8 and the frame size to $128 \times 256$. Training takes approximately 30 hours on 8 Nvidia RTX 3090 GPUs. During inference, we set the EDM denoising step to 25, the guidance scale to 1.5, and use a higher frame size of $256 \times 512$ to ensure more accurate pixel-level predictions. We cover more implementation details in the appendix.

\textbf{Datasets.} Since amodal mask can be reliably annotated only in synthetic datasets or game engines, our model is primarily trained and evaluated on synthetic datasets. We include a zero-shot evaluation on a real-world dataset to assess its generalization ability. Among synthetic datasets, \textbf{SAIL-VOS}~\cite{hu2019sailvos} includes 210 long video sequences with 162 common object classes generated from the photo-realistic game GTA-V, featuring frequent and significant occlusions. We use PySceneDetect~\cite{pyscenedetect} to identify shot transitions within these long videos, selecting only continuous scenes and segmenting them into 21,237 25-frame object sequences. \textbf{MOVi-B} and \textbf{MOVi-D}, generated by Kubrics~\cite{greff2022kubric}, feature rich annotations of simulated environments, rigid objects, and camera motions. These datasets have been adapted as video amodal segmentation benchmarks by previous studies~\cite{gao2023c2fseg,fan2023eoras} and contain 13,997 and 12,010 sequences, each with an approximate length of 25 frames. For real-world evaluation, we use \textbf{TAO-Amodal}~\cite{hsieh2023taoamodal}, a high-quality amodal tracking dataset comprising 993 video sequences in its validation set. Unlike synthetic datasets, TAO-Amodal provides only amodal bounding box annotations, as annotating amodal masks by humans is challenging. Similar to SAIL-VOS, we segment these videos into 1,392 object sequences.

\textbf{Baselines.}
We compare our method against recent baselines for both image and video amodal segmentation. For image-based amodal segmentation, our baselines include creating a convex hull around a given modal mask \cite{zhan2020pcnet}, AISFormer~\cite{tran2022aisformer}, PCNet-M~\cite{zhan2020pcnet}, and pix2gestalt~\cite{ozguroglu2024pix2gestalt}. For video-based amodal segmentation, we evaluate against SaVos~\cite{yao2022savos}, Bi-LSTM \cite{fan2023eoras, graves2012long}, EoRaS~\cite{fan2023eoras}, and C2F-Seg~\cite{gao2023c2fseg}. We discuss more details about these baselines in the appendix. Additionally, to benchmark against regression approaches, we include transformer-based VideoMAE~\cite{tong2022videomae} and SVD’s backbone 3D U-Net. We also evaluate the ground-truth modal masks.

\begin{table}
  \centering
  \caption{\textbf{Quantitative comparison on SAIL-VOS and TAO-Amodal.}  We compare our method with image-based methods (top) and video-based methods (bottom). Our method outperforms all methods on the synthetic SAIL-VOS dataset, achieving nearly a 13\% improvement in Top-1 mIoU$_{occ}$. Additionally, when trained on SAIL-VOS, our method demonstrates strong generalization, outperforming others in zero-shot evaluations on the real-world TAO-Amodal dataset. Bold values indicate the best method, and underlined values indicate the second best.}
  \footnotesize
  \begin{tabular}{@{}lccccc@{}}
    \toprule
    \multirow{2}{*}{Method} & \multicolumn{2}{c}{SAIL-VOS} & \multicolumn{3}{c}{TAO-Amodal} \\
    & mIoU & mIoU$_{occ}$ & AP$_{25}$ & AP$_{50}$ & AP$_{75}$ \\
    \midrule
    Modal & 67.89 & - & 93.73 & 82.22 & 63.12 \\
    Convex~\cite{zhan2020pcnet} & 63.18 & 27.54 & 93.73 & 82.22 & 63.12 \\
    Convex$^R$~\cite{zhan2020pcnet} & 71.21 & 34.27 & 93.73 & 82.22 & 63.12 \\
    PCNet-M~\cite{zhan2020pcnet} & 74.2 & 42.52 & 94.89 & 85.11 & 65.97 \\
    AISFormer~\cite{tran2022aisformer} & 73.51 & 39.16 & 95.45 & 81.93 & 59.84 \\
    pix2gestalt (Top-1)~\cite{ozguroglu2024pix2gestalt} & 54.83 & 26.59 & 80.73 & 57.50 & 28.95 \\
    pix2gestalt (Top-3)~\cite{ozguroglu2024pix2gestalt} & 60.79 & 33.76 & 91.80 & 71.19 & 38.80 \\
    
    \midrule
    VideoMAE~\cite{tong2022videomae} & 69.67 & 29.39 & 69.14 & 56.71 & 41.19 \\
    3D-UNet & 72.79 & 39.54 & 94.59 & 83.83 & 64.33   \\
    Ours (Top-1) & \underline{77.07} & \underline{55.12} & \underline{97.28} & \underline{89.25} & \underline{71.99} \\
    Ours (Top-3) & \textbf{79.23} & \textbf{59.69} & \textbf{98.31} & \textbf{92.46} & \textbf{77.48} \\

    \bottomrule
  \end{tabular}
  \label{tab:sailvos}
\end{table}

\begin{figure}[t]
  \centering
   \includegraphics[width=1.0\linewidth]{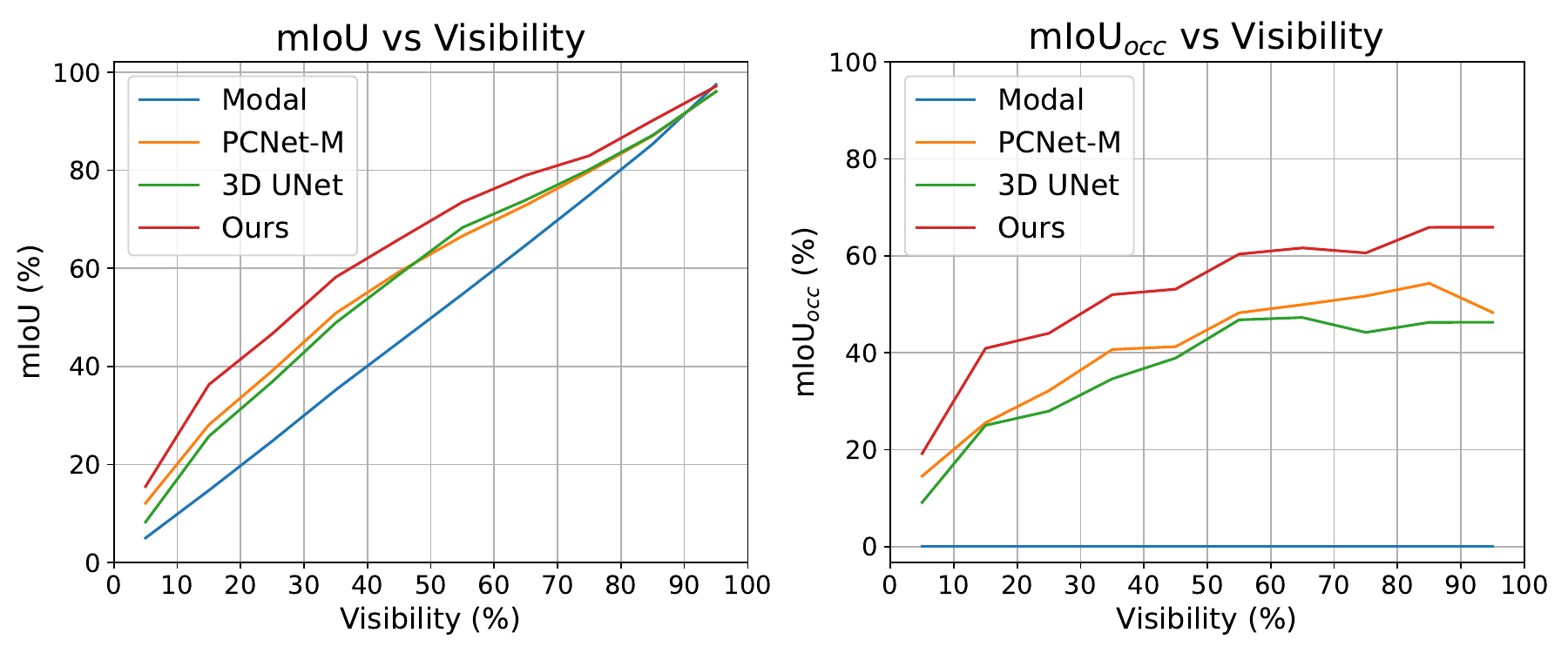}

   \caption{\textbf{Comparison across visibility levels} on SAIL-VOS. Our method outperforms the second-best image and video amodal segmentation methods across all visibility ranges (we use Top-1 metrics). This highlights the ability of our method to handle heavy occlusions, \textit{and} understand when an object is not occluded.}
   \label{fig:visibility}
\end{figure}

\begin{table}
  \centering
  \caption{\textbf{Quantitative Comparison on MOVi-B/D.} Due to strong camera motion and higher occlusions in these datasets, multi-frame methods generally outperform single-frame methods. Our method surpasses all prior state-of-the-art, achieving over a 4\% improvement in Top-1 mIoU$_{occ}$ across both datasets.} 
  \small
  \begin{tabular}{@{}lcccccc@{}}
    \toprule
    \multirow{2}{*}{Method} & \multicolumn{2}{c}{MOVi-B} &  \multicolumn{2}{c}{MOVi-D} \\
    & mIoU & mIoU$_{occ}$ & mIoU & mIoU$_{occ}$ \\
    \midrule
    Modal & 59.19 & - & 56.92 & - \\
    Convex~\cite{zhan2020pcnet} & 64.21 & 18.42 & 60.18 & 16.48 \\
    PCNet-M~\cite{zhan2020pcnet} & 65.79 & 24.02 & 64.35 & 27.31 \\
    AISFormer~\cite{tran2022aisformer} & 77.34 & 43.53 & 67.72 & 33.65 \\
    \midrule
    SaVos~\cite{yao2022savos} & 70.72 & 33.61 & 60.61 & 22.64 \\
    Bi-LSTM~\cite{graves2012long, fan2023eoras} & 77.93 & 46.21 & 68.43 & 36.00 \\
    EoRaS~\cite{fan2023eoras} & 81.76 & 49.39 & 74.1 & 38.33 \\
    C2F-Seg~\cite{gao2023c2fseg}  & - & - & 71.67 & 36.13  \\
    VideoMAE~\cite{tong2022videomae} & 78.74 & 42.86 & 70.93 & 32.78 \\
    3D-UNet & 82.16 & 49.81 & 75.65 & 40.86 \\
    Ours (Top-1) & \underline{83.51} & \underline{53.75} & \underline{77.03} & \underline{44.23} \\
    Ours (Top-3) & \textbf{83.93} & \textbf{54.56} & \textbf{77.76} & \textbf{45.6} \\
    \bottomrule
  \end{tabular}
  \label{tab:movi}
\end{table}

\textbf{Metrics.} 
Following common practice in amodal segmentation~\cite{fan2023eoras, yao2022savos, gao2023c2fseg}, we use mIoU and mIoU$_{occ}$ as evaluation metrics. Given a modal-amodal sequence pair in each frame, where the ground-truth modal mask is $\mathcal{M}_i$, and the predicted and ground-truth amodal masks are $\hat{\mathcal{A}_i}$ and $\mathcal{A}_i$, respectively, we define IoU as $\frac{\hat{\mathcal{A}_i} \cap \mathcal{A}_i}{\hat{\mathcal{A}_i} \cup \mathcal{A}_i}$ and mIoU$_{occ}$ as $\frac{(\hat{\mathcal{A}_i}-\mathcal{M}_i)  \cap (\mathcal{A}_i-\mathcal{M}_i)}{(\hat{\mathcal{A}_i}-\mathcal{M}_i) \cup (\mathcal{A}_i-\mathcal{M}_i)}$. We report the mean values across all frames in the dataset as mIoU and mIoU$_{occ}$. For TAO-Amodal, which uses bounding box evaluation instead of masks, we adopt average precision metrics used in a recent amodal tracking work~\cite{hsieh2023taoamodal} -- AP$_{25}$, AP$_{50}$, and AP$_{75}$, based on varying IoU thresholds calculated over bounding box areas. Additionally, to account for the multimodal generation capability of diffusion-based methods, we adopt a probabilistic evaluation with Top-K metrics~\cite{khurana2021detecting}, selecting the best IoU or AP score in each frame from $K$ predictions.

\subsection{Comparison to state-of-the-art}
Table~\ref{tab:sailvos} shows the quantitative comparisons on SAIL-VOS and TAO-Amodal, where our method surpasses all baselines. Notably, it achieves nearly 13\% improvement over the second-best method, PCNet-M~\cite{zhan2020pcnet}, in terms of mIoU$_{occ}$, highlighting effective completion of occluded object regions. Despite being trained exclusively on synthetic SAIL-VOS, a zero-shot evaluation on TAO-Amodal highlights the strong generalization of our model. We posit that, in addition to leveraging foundational knowledge and rich priors from the large-scale pretraining of SVD, our model is able to learn temporal cues that help it amodally complete any unseen object classes from neighboring frames. Figure~\ref{fig:visibility} further illustrates our method's consistent performance across all visibility ranges on SAIL-VOS~\cite{hu2019sailvos}, indicating that our method can realistically hallucinate masks in occluded regions, across the entire range of visibility levels.

\begin{figure}[t]
  \centering
   \includegraphics[width=1.0\linewidth]{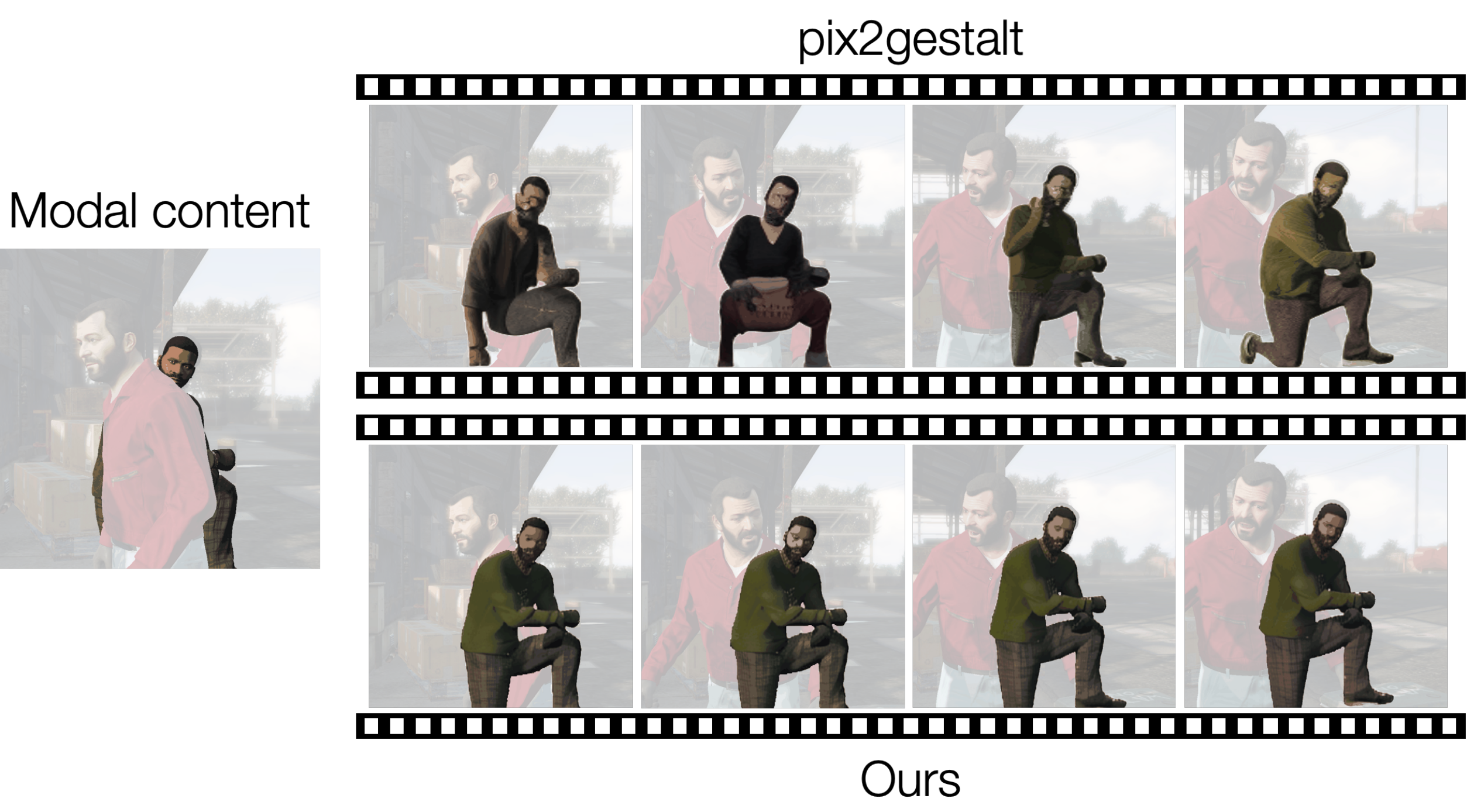}

   \caption{\textbf{Temporal consistency comparison} with an image amodal segmentation method. We highlight the lack of temporal coherence in a single-frame diffusion based method, pix2gestalt, for both the predicted amodal segmentation mask and the RGB content for the occluded person in the example shown. By leveraging temporal priors, our approach achieves significantly higher temporal consistency across occlusions.}
   \label{fig:temporal_consist}
\end{figure}

We also compare with a single-frame diffusion-based method, pix2gestalt \cite{ozguroglu2024pix2gestalt}, in Table~\ref{tab:sailvos}. 
Perhaps unsurprisingly, pix2gestalt performs poorly on these video benchmarks, likely because many objects undergo \textit{high} degrees of occlusion. Since, pix2gestalt is a single-frame method, we also find that its predictions vary significantly across frames and lack temporal coherence (c.f. Figure~\ref{fig:temporal_consist}). 
In contrast, our method does drastically better because it can handle both, high occlusions and temporal coherence across frames.

Table~\ref{tab:movi} provides quantitative comparisons on the MOVi-B/D datasets, where our method beats the prior state-of-the-art. Despite strong camera motion in MOVi-B/D, our model adapts well \textit{without} access to camera extrinsics or optical flow (unlike some baselines~\cite{yao2022savos, fan2023eoras}). We posit that our method is able to use the 3D priors from Stable Video Diffusion and is therefore, successfully able to maintain consistent object shapes from different view-points. Notably, prior works on MOVi-B/D are evaluated using a cropped modal bounding box enlarged by 2 times as input; we adopt the same setting here for a fair comparison. However, we observed that using the full, uncropped image as input can significantly enhance model performance, and we include these results in the appendix.

For content completion, due to the lack of ground truth and standardized metrics, we conducted a user study on 20 randomly selected sequences from SAIL-VOS and TAO-Amodal. In this user study, we did A/B testing and forced participants to choose between our method and pix2gestalt. We found that users showed a preference of 85.6\% for our method over pix2gestalt.

\begin{figure*}[htbp]
  \centering
   \includegraphics[width=1.0\linewidth]{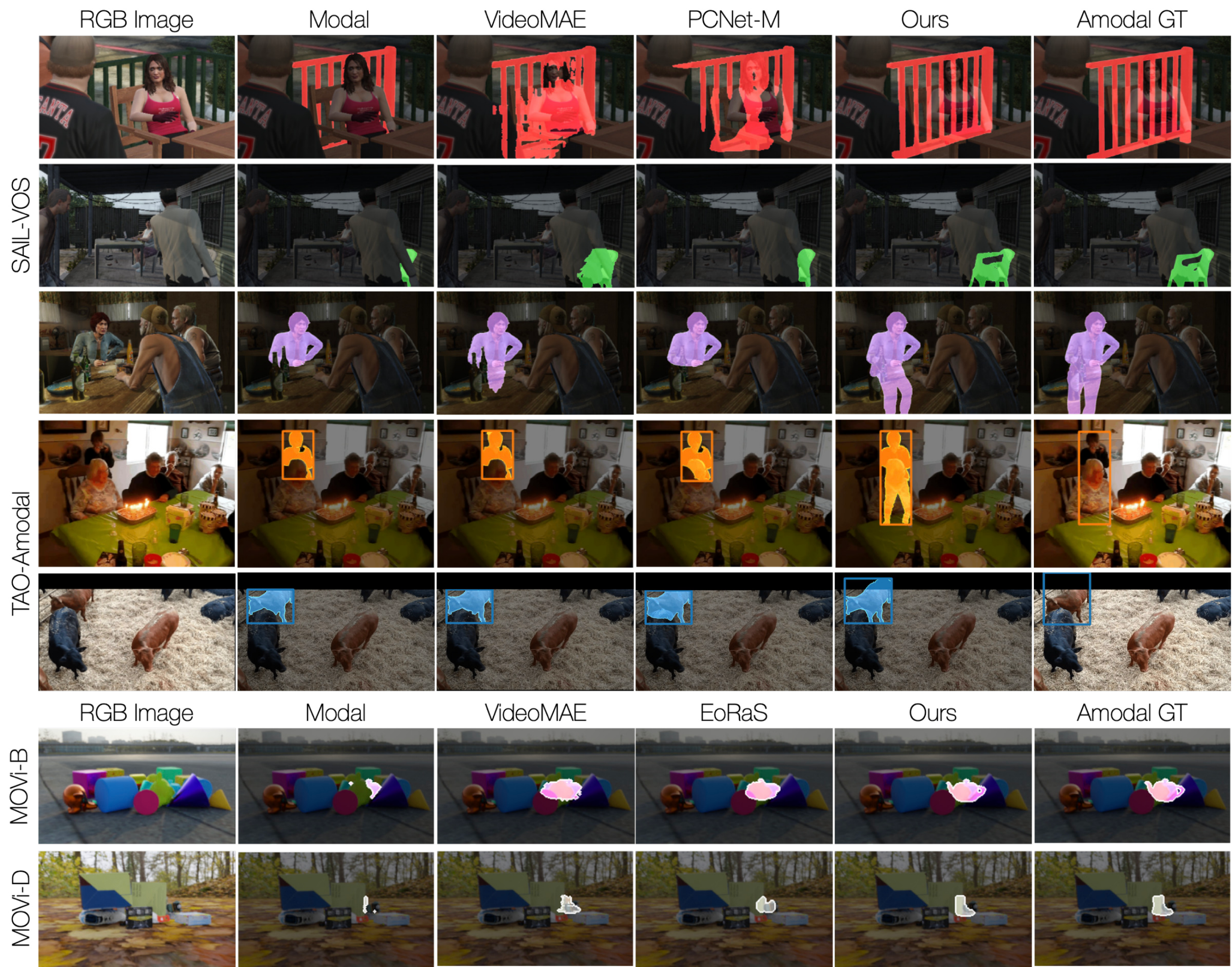}

   \caption{\textbf{Qualitative comparison} of amodal segmentation methods across diverse datasets. Our method leverages strong shape priors, such as for humans, chairs, and teapots, to generate clean and realistic object shapes. It also excels in handling heavy occlusions; even when objects are nearly fully occluded (e.g., ``chair" in the second row of SAIL-VOS), our method achieves high-fidelity shape completion by utilizing temporal priors. Note that TAO-Amodal contains out-of-frame occlusions which none of the methods are trained for, but our method is able to handle such cases.}
   \label{fig:sailvos_comp}
\end{figure*}

\begin{figure*}[htbp]
  \centering
   \includegraphics[width=1.0\linewidth]{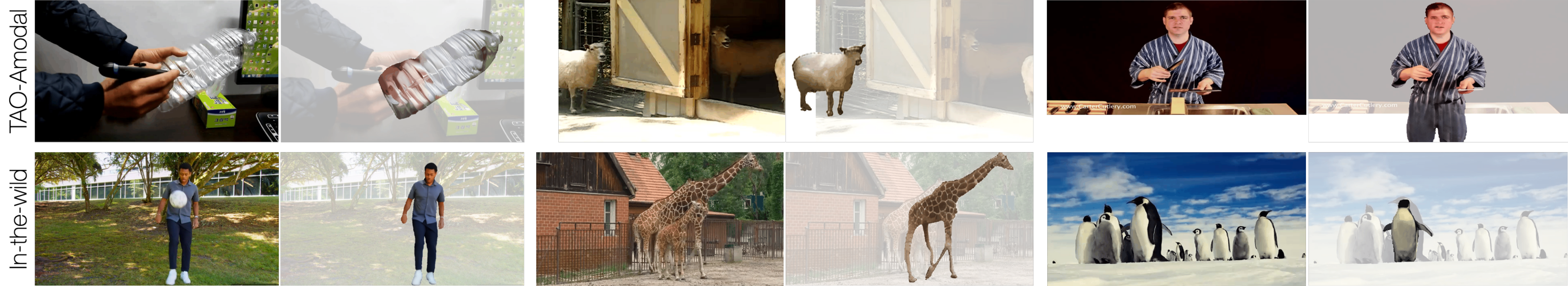}

   \caption{\textbf{Qualitative results for content completion.} Although our content completion module, initialized from pretrained SVD weights, is finetuned solely on synthetic SAIL-VOS, it achieves photorealistic, high-fidelity object inpainting even in real-world scenarios. Furthermore, our method can complete \textit{unseen} categories, such as giraffes and plastic bottle, likely due to its ability to transfer styles and patterns from the visible parts of objects to occluded areas in the current or neighboring frames. We show examples from TAO-Amodal (top) and in-the-wild YouTube videos (bottom).}
   \label{fig:content_comp}
\end{figure*}

\subsection{Ablation studies}
\textbf{Conditioning.}
Here we ablate our choice of modal mask and pseudo-depth conditioning. We also examine the effect of additionally using RGB video frames as conditioning. As shown in Table~\ref{tab:ablation_condition}, adding RGB or pseudo-depth information improves model performance, with pseudo-depth providing a more substantial enhancement. Although combining both RGB and pseudo-depth yields a higher mIoU on SAIL-VOS, conditioning on pseudo-depth alone outperforms across other metrics. This supports our claim that pseudo-depth is a more generalizable modality, and the dependence on texture and appearance cues in fact hinders generalization of our model to TAO-Amodal.

\textbf{Training strategies.} 
Table~\ref{tab:ablation_training_str} shows the impact of our two-stage fine-tuning strategy and use of zero convolutions. Compared to randomly initializing the new input convolution layer in the 3D U-Net, we find that zero convolution significantly improves model performance. Additionally, compared to training with both modal mask and pseudo-depth conditionings from scratch, two-stage fine-tuning (where we train with modal masks first and then add pseudo-depth) leads to further quantitative improvements. We also include an ablation study on the weights initialization in the appendix.

\begin{table}
  \centering
  \caption{\textbf{Ablation study for input conditioning.} We study the effect of conditioning our model on different input modalities. Results show that pseudo-depth conditioning yields greater performance improvements than RGB conditioning across almost all metrics. We therefore drop RGB conditioning in the final method.}
  \footnotesize
  \begin{tabular}{@{}cccccccc@{}}
    \toprule
    \multicolumn{3}{c}{Conditions} & \multicolumn{2}{c}{SAIL-VOS} & \multicolumn{3}{c}{TAO-Amodal}\\
    mask & RGB & depth & mIoU & mIoU$_{occ}$ & AP$_{25}$ & AP$_{50}$ & AP$_{75}$ \\
    \midrule
    \ding{51} & \ding{55} & \ding{55} & 75.17 & 51.28 & 94.89 & 85.03 & 66.87 \\
    \ding{51} & \ding{51} & \ding{55} & 76.59 & 53.3 & 95.86 & 86.59 & 70.12 \\
    \ding{51} & \ding{55} & \ding{51} & 77.07 & \textbf{55.12} & \textbf{97.28} & \textbf{89.25} & \textbf{69.65} \\
    \ding{51} & \ding{51} & \ding{51} & \textbf{77.19} & 54.59 & 96.6 & 87.16 & 69.64 \\
    \bottomrule
  \end{tabular}
  \label{tab:ablation_condition}
\end{table}

\begin{table}
  \centering
  \caption{\textbf{Ablation study for training strategies.} We study the effect of two-stage finetuning for segmentation. We find that zero convolution helps significantly, while two-stage fine-tuning gives us an additional, moderate improvement.}
  \footnotesize
  \begin{tabular}{@{}ccccccc@{}}
    \toprule
    \multicolumn{2}{c}{Training strategies} & \multicolumn{2}{c}{SAIL-VOS} & \multicolumn{3}{c}{TAO-Amodal}\\
    2-stage ft. & zero-conv & mIoU & mIoU$_{occ}$ & AP$_{25}$ & AP$_{50}$ & AP$_{75}$ \\
    \midrule
    \ding{55} & \ding{55}  & 73.73 & 41.35 & 96.27 & 85.93 & 66.45 \\
    \ding{51} & \ding{55}  & 72.72 & 32.23 & 95.38 & 86.1 & 68.74 \\
    \ding{55} & \ding{51}  & 76.92 & 54.25 & 96.58 & 87.64 & 69.34 \\
    \ding{51} & \ding{51} & \textbf{77.07} & \textbf{55.12} & \textbf{97.28} & \textbf{89.25} & \textbf{71.99}\\
    \bottomrule
  \end{tabular}
  \label{tab:ablation_training_str}
\end{table}

\begin{figure}[t]
  \centering
   \includegraphics[width=1.0\linewidth]{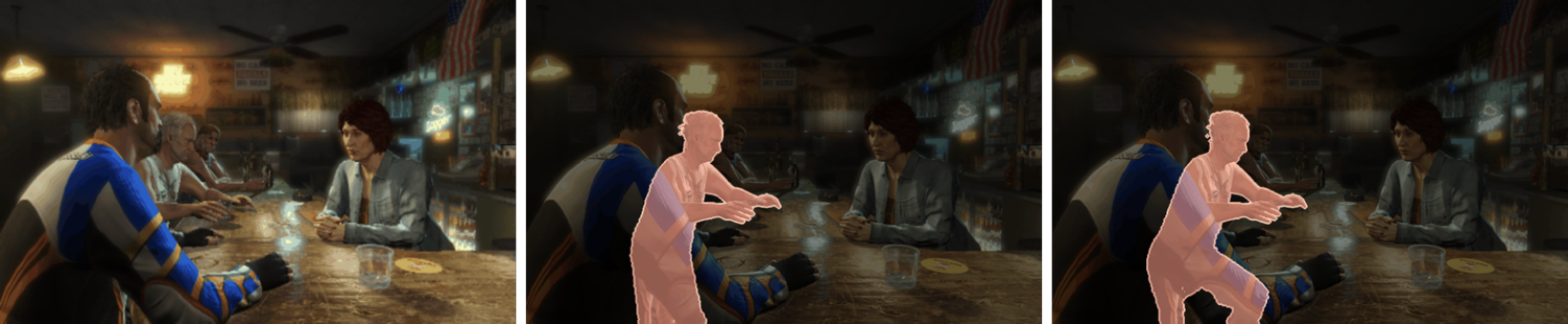}

   \caption{We show an example of \textbf{multi-modal generation} from our diffusion model. Since there are multiple plausible explanations for the shape of the person in his occluded region, our model predicts two such plausible amodal masks (with the person's occluded legs in two different orientations). 
   }
   \label{fig:multimodal}
\end{figure}

\begin{figure}[t]
  \centering
   \includegraphics[width=1.0\linewidth]{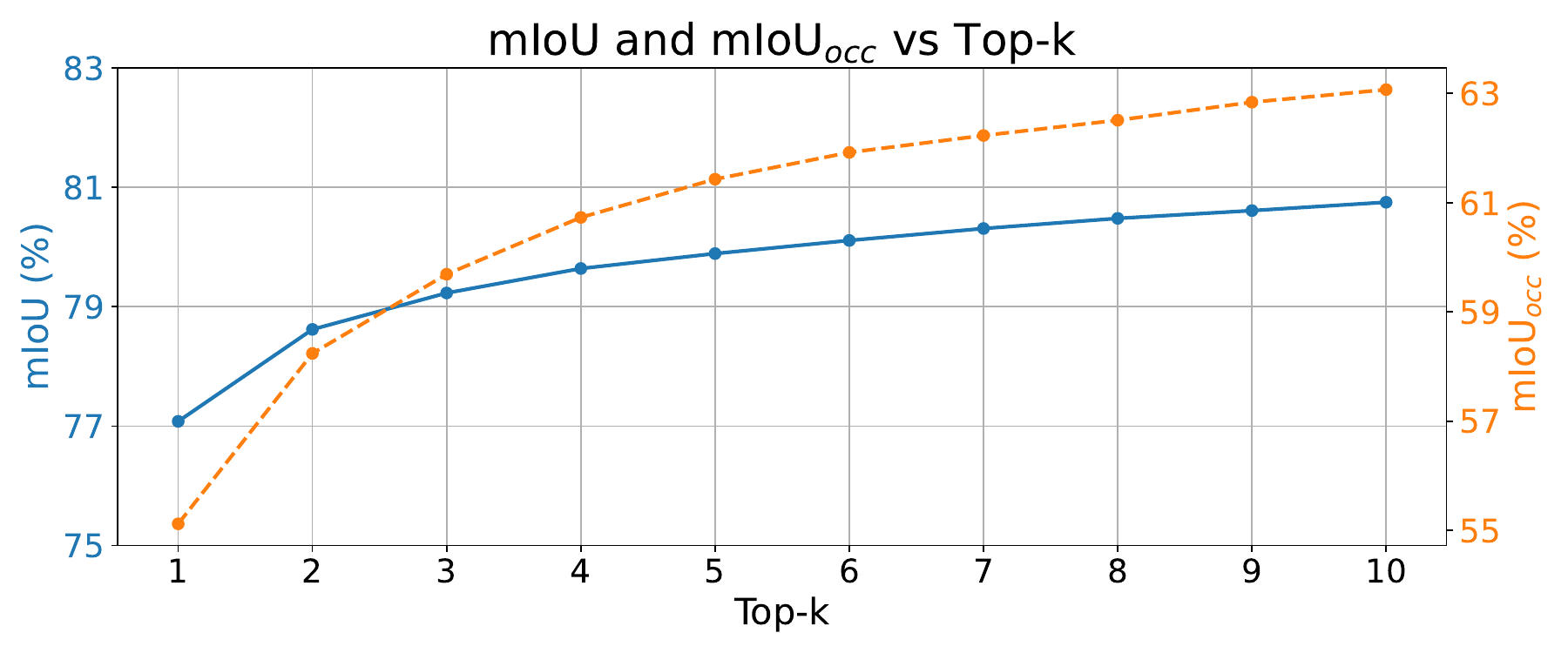}

   \caption{\textbf{Ablation of Top-K} on SAIL-VOS. We find that increasing the number of output samples from our method, $K$, leads to improvements in both mIoU and mIoU$_{occ}$; however, these improvements gradually plateau as we increase $K$.}
   \label{fig:ablation_topk}
\end{figure}

\textbf{Top-k evaluation.} 
Like other diffusion models, ours also supports multimodal generation. For instance, when parts of an object remain consistently occluded in a video (e.g., a person’s legs), multiple plausible interpretations of the occluded area (e.g., standing, sitting) may exist, as illustrated in Figure~\ref{fig:multimodal}. By setting different random seeds, our model generates varying predictions for the same input, leading to different IoU values against the ground truth. Figure~\ref{fig:ablation_topk} reports the Top-10 mIoU and mIoU$_{occ}$ results for our model. As expected, performance improves with the number of outputs, although the gains gradually diminish.

\subsection{Applications}
Our amodal segmentation masks and content completions can enable multiple downstream applications. We touch on three such applications in the appendix -- 4D reconstruction, scene manipulation, and pseudo-groundtruth generation. First, we find that one can use monocular video to multi-view generation methods like SV4D \cite{xie2024sv4d} to reconstruct dynamic objects across space and time, even when they get fully occluded. Second, we show that once all objects in a scene are de-occluded, they can be rearranged in the scene to simulate different object interactions and realities. Third, we can generate amodal segmentation pseudo-groundtruth on real-world datasets to fill in the gap for lack of real-world training data for video amodal segmentation. Please see appendix for more details and visual examples. %

\section{Discussion}

In this work, we focus on the problem of video amodal segmentation -- segmenting objects to their full extent even when they may be partially or fully occluded in videos. We lean on the large-scale pretraining of video foundation models and adapt Stable Video Diffusion \cite{svd} for the task of video amodal segmentation. Given an object's modal mask sequence, and pseudo-depth maps of the scene, we aim to predict amodal masks of the occluded object. This amodal mask is used by another model to inpaint the RGB content in the object's occluded region. One of the key insights of our work is that one can use the shape and temporal priors learnt by video foundation models. More crucially, our multi-frame setup allows us to propagate mask and RGB content from the frames where an object maybe fully visible to the frames of high occlusion.
We find that, our models can perform exceedingly well even for unseen categories, likely because of their pretraining on foundational data.

\paragraph{Acknowledgments} We would like to thank Carl Vondrick, Ege Ozguroglu and Achal Dave for insightful discussions and clarifications on pix2gestalt's evaluation protocol. Mosam Dabhi helped with demonstrating the application of our work to 4D reconstruction.
{
    \small
    \bibliographystyle{ieeenat_fullname}
    \bibliography{main}

\begin{thebibliography}{62}
\providecommand{\natexlab}[1]{#1}
\providecommand{\url}[1]{\texttt{#1}}
\expandafter\ifx\csname urlstyle\endcsname\relax
  \providecommand{\doi}[1]{doi: #1}\else
  \providecommand{\doi}{doi: \begingroup \urlstyle{rm}\Url}\fi

\bibitem[Athar et~al.(2023)Athar, Luiten, Voigtlaender, Khurana, Dave, Leibe,
  and Ramanan]{athar2023burst}
Ali Athar, Jonathon Luiten, Paul Voigtlaender, Tarasha Khurana, Achal Dave,
  Bastian Leibe, and Deva Ramanan.
\newblock Burst: A benchmark for unifying object recognition, segmentation and
  tracking in video.
\newblock In \emph{Proceedings of the IEEE/CVF Winter Conference on
  Applications of Computer Vision}, pages 1674--1683, 2023.

\bibitem[Baillargeon and DeVos(1991)]{baillargeon1991object}
Renée Baillargeon and Julie DeVos.
\newblock Object permanence in young infants: Further evidence.
\newblock \emph{Child Development}, 62\penalty0 (6):\penalty0 1227--1246, 1991.

\bibitem[Blattmann et~al.(2023{\natexlab{a}})Blattmann, Dockhorn, Kulal,
  Mendelevitch, Kilian, Lorenz, Levi, et~al.]{svd}
Andreas Blattmann, Tim Dockhorn, Sumith Kulal, Daniel Mendelevitch, Maciej
  Kilian, Dominik Lorenz, Yam Levi, et~al.
\newblock Stable video diffusion: Scaling latent video diffusion models to
  large datasets.
\newblock \emph{arXiv preprint arXiv:2311.15127}, 2023{\natexlab{a}}.

\bibitem[Blattmann et~al.(2023{\natexlab{b}})Blattmann, Rombach, Ling,
  Dockhorn, Kim, Fidler, and Kreis]{vldm}
Andreas Blattmann, Robin Rombach, Huan Ling, Tim Dockhorn, Seung~Wook Kim,
  Sanja Fidler, and Karsten Kreis.
\newblock Align your latents: High-resolution video synthesis with latent
  diffusion models.
\newblock In \emph{CVPR}, pages 22563--22575, 2023{\natexlab{b}}.

\bibitem[{Breakthrough}(2024)]{pyscenedetect}
{Breakthrough}.
\newblock {PySceneDetect}: Video scene cut detection tool, 2024.
\newblock Accessed: 2024-11-12.

\bibitem[Brooks et~al.(2023)Brooks, Holynski, and
  Efros]{brooks2023instructpix2pix}
Tim Brooks, Aleksander Holynski, and Alexei~A. Efros.
\newblock Instructpix2pix: Learning to follow image editing instructions.
\newblock In \emph{Proceedings of the IEEE/CVF Conference on Computer Vision
  and Pattern Recognition}, pages 18392--18402, 2023.

\bibitem[Dave et~al.(2020)Dave, Khurana, Tokmakov, Schmid, and
  Ramanan]{dave2020tao}
Achal Dave, Tarasha Khurana, Pavel Tokmakov, Cordelia Schmid, and Deva Ramanan.
\newblock Tao: A large-scale benchmark for tracking any object.
\newblock In \emph{Computer Vision--ECCV 2020: 16th European Conference,
  Glasgow, UK, August 23--28, 2020, Proceedings, Part V}, pages 436--454.
  Springer International Publishing, 2020.

\bibitem[Ehsani et~al.(2018)Ehsani, Mottaghi, and Farhadi]{ehsani2018segan}
Kiana Ehsani, Roozbeh Mottaghi, and Ali Farhadi.
\newblock Segan: Segmenting and generating the invisible.
\newblock In \emph{Proceedings of the IEEE Conference on Computer Vision and
  Pattern Recognition}, pages 6144--6153, 2018.

\bibitem[Fan et~al.(2023)Fan, Lei, Qian, Yu, Xiao, He, Zhang, and
  Fu]{fan2023eoras}
Ke Fan, Jingshi Lei, Xuelin Qian, Miaopeng Yu, Tianjun Xiao, Tong He, Zheng
  Zhang, and Yanwei Fu.
\newblock Rethinking amodal video segmentation from learning supervised signals
  with object-centric representation.
\newblock In \emph{Proceedings of the IEEE/CVF International Conference on
  Computer Vision}, pages 1272--1281, 2023.

\bibitem[Follmann et~al.(2019)Follmann, König, Härtinger, Klostermann, and
  Böttger]{follmann2019orcnn}
Patrick Follmann, Rebecca König, Philipp Härtinger, Michael Klostermann, and
  Tobias Böttger.
\newblock Learning to see the invisible: End-to-end trainable amodal instance
  segmentation.
\newblock In \emph{2019 IEEE Winter Conference on Applications of Computer
  Vision (WACV)}, pages 1328--1336. IEEE, 2019.

\bibitem[Gao et~al.(2023)Gao, Qian, Wang, Xiao, He, Zhang, and
  Fu]{gao2023c2fseg}
Jianxiong Gao, Xuelin Qian, Yikai Wang, Tianjun Xiao, Tong He, Zheng Zhang, and
  Yanwei Fu.
\newblock Coarse-to-fine amodal segmentation with shape prior.
\newblock In \emph{Proceedings of the IEEE/CVF International Conference on
  Computer Vision}, pages 1262--1271, 2023.

\bibitem[Geiger et~al.(2013)Geiger, Lenz, Stiller, and
  Urtasun]{geiger2013kitti}
Andreas Geiger, Philip Lenz, Christoph Stiller, and Raquel Urtasun.
\newblock Vision meets robotics: The kitti dataset.
\newblock \emph{The International Journal of Robotics Research}, 32\penalty0
  (11):\penalty0 1231--1237, 2013.

\bibitem[Graves(2012)]{graves2012long}
Alex Graves.
\newblock \emph{Long short-term memory}, pages 37--45.
\newblock 2012.

\bibitem[Greff et~al.(2022)Greff, Belletti, Beyer, Doersch, Du, Duckworth,
  Fleet, et~al.]{greff2022kubric}
Klaus Greff, Francois Belletti, Lucas Beyer, Carl Doersch, Yilun Du, Daniel
  Duckworth, David~J. Fleet, et~al.
\newblock Kubric: A scalable dataset generator.
\newblock In \emph{Proceedings of the IEEE/CVF Conference on Computer Vision
  and Pattern Recognition}, pages 3749--3761, 2022.

\bibitem[Ho and Salimans(2022)]{cfg}
Jonathan Ho and Tim Salimans.
\newblock Classifier-free diffusion guidance.
\newblock \emph{arXiv preprint arXiv:2207.12598}, 2022.

\bibitem[Ho et~al.(2020)Ho, Jain, and Abbeel]{ddpm}
Jonathan Ho, Ajay Jain, and Pieter Abbeel.
\newblock Denoising diffusion probabilistic models.
\newblock In \emph{NeurIPS}, pages 6840--6851, 2020.

\bibitem[Ho et~al.(2022)Ho, Chan, Saharia, Whang, Gao, Gritsenko, Kingma,
  et~al.]{ho2022imagenv}
Jonathan Ho, William Chan, Chitwan Saharia, Jay Whang, Ruiqi Gao, Alexey
  Gritsenko, Diederik~P. Kingma, et~al.
\newblock Imagen video: High definition video generation with diffusion models.
\newblock \emph{arXiv preprint arXiv:2210.02303}, 2022.

\bibitem[Hsieh et~al.(2023)Hsieh, Khurana, Dave, and
  Ramanan]{hsieh2023taoamodal}
Cheng-Yen Hsieh, Tarasha Khurana, Achal Dave, and Deva Ramanan.
\newblock Tao-amodal: A benchmark for tracking any object amodally.
\newblock \emph{arXiv preprint arXiv:2312.12433}, 2023.

\bibitem[Hu et~al.(2019)Hu, Chen, Hui, Huang, and Schwing]{hu2019sailvos}
Yuan-Ting Hu, Hong-Shuo Chen, Kexin Hui, Jia-Bin Huang, and Alexander~G.
  Schwing.
\newblock Sail-vos: Semantic amodal instance level video object segmentation-a
  synthetic dataset and baselines.
\newblock In \emph{Proceedings of the IEEE/CVF Conference on Computer Vision
  and Pattern Recognition}, pages 3105--3115, 2019.

\bibitem[Karras et~al.(2022)Karras, Aittala, Aila, and Laine]{edm}
Tero Karras, Miika Aittala, Timo Aila, and Samuli Laine.
\newblock Elucidating the design space of diffusion-based generative models.
\newblock In \emph{NeurIPS}, pages 26565--26577, 2022.

\bibitem[Kavsek(2004)]{kavsek2004influence}
Michael Kavsek.
\newblock The influence of context on amodal completion in 5-and 7-month-old
  infants.
\newblock \emph{Journal of Cognition and Development}, 5\penalty0 (2):\penalty0
  159--184, 2004.

\bibitem[Ke et~al.(2024)Ke, Obukhov, Huang, Metzger, Daudt, and
  Schindler]{ke2024marigold}
Bingxin Ke, Anton Obukhov, Shengyu Huang, Nando Metzger, Rodrigo~Caye Daudt,
  and Konrad Schindler.
\newblock Repurposing diffusion-based image generators for monocular depth
  estimation.
\newblock In \emph{Proceedings of the IEEE/CVF Conference on Computer Vision
  and Pattern Recognition}, pages 9492--9502, 2024.

\bibitem[Ke et~al.(2021)Ke, Tai, and Tang]{ke2021bcnet}
Lei Ke, Yu-Wing Tai, and Chi-Keung Tang.
\newblock Deep occlusion-aware instance segmentation with overlapping bilayers.
\newblock In \emph{Proceedings of the IEEE/CVF Conference on Computer Vision
  and Pattern Recognition}, pages 4019--4028, 2021.

\bibitem[Khurana et~al.(2021)Khurana, Dave, and Ramanan]{khurana2021detecting}
Tarasha Khurana, Achal Dave, and Deva Ramanan.
\newblock Detecting invisible people.
\newblock In \emph{Proceedings of the IEEE/CVF International Conference on
  Computer Vision}, pages 3174--3184, 2021.

\bibitem[Kingma(2013)]{kingma2013vae}
Diederik~P. Kingma.
\newblock Auto-encoding variational bayes.
\newblock \emph{arXiv preprint arXiv:1312.6114}, 2013.

\bibitem[Kirillov et~al.(2023)Kirillov, Mintun, Ravi, Mao, Rolland, Gustafson,
  Xiao, et~al.]{kirillov2023sam}
Alexander Kirillov, Eric Mintun, Nikhila Ravi, Hanzi Mao, Chloe Rolland, Laura
  Gustafson, Tete Xiao, et~al.
\newblock Segment anything.
\newblock In \emph{Proceedings of the IEEE/CVF International Conference on
  Computer Vision}, pages 4015--4026, 2023.

\bibitem[Koffka(2013)]{koffka2013gestalt}
Kurt Koffka.
\newblock \emph{Principles of Gestalt Psychology}.
\newblock Routledge, 2013.

\bibitem[Li and Malik(2016)]{li2016amodal}
Ke Li and Jitendra Malik.
\newblock Amodal instance segmentation.
\newblock In \emph{Computer Vision--ECCV 2016: 14th European Conference,
  Amsterdam, The Netherlands, October 11-14, 2016, Proceedings, Part II}, pages
  677--693. Springer International Publishing, 2016.

\bibitem[Ling et~al.(2020)Ling, Acuna, Kreis, Kim, and
  Fidler]{ling2020amodalvae}
Huan Ling, David Acuna, Karsten Kreis, Seung~Wook Kim, and Sanja Fidler.
\newblock Variational amodal object completion.
\newblock \emph{Advances in Neural Information Processing Systems},
  33:\penalty0 16246--16257, 2020.

\bibitem[Liu et~al.(2024)Liu, Sun, Wang, Wang, Sun, Ye, Zhang, and
  Duan]{liu2024reconx}
Fangfu Liu, Wenqiang Sun, Hanyang Wang, Yikai Wang, Haowen Sun, Junliang Ye,
  Jun Zhang, and Yueqi Duan.
\newblock Reconx: Reconstruct any scene from sparse views with video diffusion
  model.
\newblock \emph{arXiv preprint arXiv:2408.16767}, 2024.

\bibitem[Liu et~al.(2023)Liu, Wu, Hoorick, Tokmakov, Zakharov, and
  Vondrick]{liu2023zero}
Ruoshi Liu, Rundi Wu, Basile~Van Hoorick, Pavel Tokmakov, Sergey Zakharov, and
  Carl Vondrick.
\newblock Zero-1-to-3: Zero-shot one image to 3d object.
\newblock In \emph{Proceedings of the IEEE/CVF International Conference on
  Computer Vision}, pages 9298--9309, 2023.

\bibitem[Lugmayr et~al.(2022)Lugmayr, Danelljan, Romero, Yu, Timofte, and
  Gool]{lugmayr2022repaint}
Andreas Lugmayr, Martin Danelljan, Andres Romero, Fisher Yu, Radu Timofte, and
  Luc~Van Gool.
\newblock Repaint: Inpainting using denoising diffusion probabilistic models.
\newblock In \emph{Proceedings of the IEEE/CVF Conference on Computer Vision
  and Pattern Recognition}, pages 11461--11471, 2022.

\bibitem[Martin et~al.(2001)Martin, Fowlkes, Tal, and
  Malik]{martin2001database}
David Martin, Charless Fowlkes, Doron Tal, and Jitendra Malik.
\newblock A database of human segmented natural images and its application to
  evaluating segmentation algorithms and measuring ecological statistics.
\newblock In \emph{Proceedings of the Eighth IEEE International Conference on
  Computer Vision (ICCV 2001)}, pages 416--423. IEEE, 2001.

\bibitem[Menapace et~al.(2024)Menapace, Siarohin, Skorokhodov, Deyneka, Chen,
  Kag, Fang, et~al.]{snapv}
Willi Menapace, Aliaksandr Siarohin, Ivan Skorokhodov, Ekaterina Deyneka,
  Tsai-Shien Chen, Anil Kag, Yuwei Fang, et~al.
\newblock Snap video: Scaled spatiotemporal transformers for text-to-video
  synthesis.
\newblock In \emph{CVPR}, pages 7038--7048, 2024.

\bibitem[Otsuka et~al.(2006)Otsuka, Kanazawa, and
  Yamaguchi]{otsuka2006development}
Yumiko Otsuka, So Kanazawa, and Masami~K. Yamaguchi.
\newblock Development of modal and amodal completion in infants.
\newblock \emph{Perception}, 35\penalty0 (9):\penalty0 1251--1264, 2006.

\bibitem[Ozguroglu et~al.(2024)Ozguroglu, Liu, Surís, Chen, Dave, Tokmakov,
  and Vondrick]{ozguroglu2024pix2gestalt}
Ege Ozguroglu, Ruoshi Liu, Dídac Surís, Dian Chen, Achal Dave, Pavel
  Tokmakov, and Carl Vondrick.
\newblock pix2gestalt: Amodal segmentation by synthesizing wholes.
\newblock In \emph{Proceedings of the IEEE/CVF Conference on Computer Vision
  and Pattern Recognition}, pages 3931--3940, 2024.

\bibitem[Podell et~al.(2023)Podell, English, Lacey, Blattmann, Dockhorn,
  Müller, Penna, and Rombach]{podell2023sdxl}
Dustin Podell, Zion English, Kyle Lacey, Andreas Blattmann, Tim Dockhorn, Jonas
  Müller, Joe Penna, and Robin Rombach.
\newblock Sdxl: Improving latent diffusion models for high-resolution image
  synthesis.
\newblock \emph{arXiv preprint arXiv:2307.01952}, 2023.

\bibitem[Qi et~al.(2019)Qi, Jiang, Liu, Shen, and Jia]{qi2019kins}
Lu Qi, Li Jiang, Shu Liu, Xiaoyong Shen, and Jiaya Jia.
\newblock Amodal instance segmentation with kins dataset.
\newblock In \emph{Proceedings of the IEEE/CVF Conference on Computer Vision
  and Pattern Recognition}, pages 3014--3023, 2019.

\bibitem[Radford et~al.(2021)Radford, Kim, Hallacy, Ramesh, Goh, Agarwal,
  Sastry, et~al.]{radford2021clip}
Alec Radford, Jong~Wook Kim, Chris Hallacy, Aditya Ramesh, Gabriel Goh,
  Sandhini Agarwal, Girish Sastry, et~al.
\newblock Learning transferable visual models from natural language
  supervision.
\newblock In \emph{International Conference on Machine Learning}, pages
  8748--8763. PMLR, 2021.

\bibitem[Ramesh et~al.(2021)Ramesh, Pavlov, Goh, Gray, Voss, Radford, Chen, and
  Sutskever]{ramesh2021dalle}
Aditya Ramesh, Mikhail Pavlov, Gabriel Goh, Scott Gray, Chelsea Voss, Alec
  Radford, Mark Chen, and Ilya Sutskever.
\newblock Zero-shot text-to-image generation.
\newblock In \emph{International Conference on Machine Learning}, pages
  8821--8831. PMLR, 2021.

\bibitem[Ravi et~al.(2024)Ravi, Gabeur, Hu, Hu, Ryali, Ma, Khedr,
  et~al.]{ravi2024sam2}
Nikhila Ravi, Valentin Gabeur, Yuan-Ting Hu, Ronghang Hu, Chaitanya Ryali,
  Tengyu Ma, Haitham Khedr, et~al.
\newblock Sam 2: Segment anything in images and videos.
\newblock \emph{arXiv preprint arXiv:2408.00714}, 2024.

\bibitem[Reddy et~al.(2022)Reddy, Tamburo, and Narasimhan]{reddy2022walt}
N.~Dinesh Reddy, Robert Tamburo, and Srinivasa~G. Narasimhan.
\newblock Walt: Watch and learn 2d amodal representation from time-lapse
  imagery.
\newblock In \emph{Proceedings of the IEEE/CVF Conference on Computer Vision
  and Pattern Recognition}, pages 9356--9366, 2022.

\bibitem[Rombach et~al.(2022)Rombach, Blattmann, Lorenz, Esser, and Ommer]{ldm}
Robin Rombach, Andreas Blattmann, Dominik Lorenz, Patrick Esser, and Björn
  Ommer.
\newblock High-resolution image synthesis with latent diffusion models.
\newblock In \emph{CVPR}, pages 10684--10695, 2022.

\bibitem[Ronneberger et~al.(2015)Ronneberger, Fischer, and
  Brox]{ronneberger2015unet}
Olaf Ronneberger, Philipp Fischer, and Thomas Brox.
\newblock U-net: Convolutional networks for biomedical image segmentation.
\newblock In \emph{Medical Image Computing and Computer-Assisted
  Intervention--MICCAI 2015: 18th International Conference, Munich, Germany,
  October 5-9, 2015, Proceedings, Part III}, pages 234--241. Springer
  International Publishing, 2015.

\bibitem[Saharia et~al.(2022)Saharia, Chan, Saxena, Li, Whang, Denton,
  Ghasemipour, et~al.]{saharia2022imagen}
Chitwan Saharia, William Chan, Saurabh Saxena, Lala Li, Jay Whang, Emily~L.
  Denton, Kamyar Ghasemipour, et~al.
\newblock Photorealistic text-to-image diffusion models with deep language
  understanding.
\newblock \emph{Advances in Neural Information Processing Systems},
  35:\penalty0 36479--36494, 2022.

\bibitem[Salimans and Ho(2022)]{vpred}
Tim Salimans and Jonathan Ho.
\newblock Progressive distillation for fast sampling of diffusion models.
\newblock \emph{arXiv preprint arXiv:2202.00512}, 2022.

\bibitem[Sargent et~al.(2023)Sargent, Li, Shah, Herrmann, Yu, Zhang, Chan,
  et~al.]{sargent2023zeronvs}
Kyle Sargent, Zizhang Li, Tanmay Shah, Charles Herrmann, Hong-Xing Yu, Yunzhi
  Zhang, Eric~Ryan Chan, et~al.
\newblock Zeronvs: Zero-shot 360-degree view synthesis from a single real
  image.
\newblock \emph{arXiv preprint arXiv:2310.17994}, 2023.

\bibitem[Song et~al.(2020)Song, Meng, and Ermon]{song2020ddim}
Jiaming Song, Chenlin Meng, and Stefano Ermon.
\newblock Denoising diffusion implicit models.
\newblock \emph{arXiv preprint arXiv:2010.02502}, 2020.

\bibitem[Tangemann et~al.(2021)Tangemann, Schneider, Kügelgen, Locatello,
  Gehler, Brox, Kümmerer, Bethge, and Schölkopf]{tangemann2021fishbowl}
Matthias Tangemann, Steffen Schneider, Julius~Von Kügelgen, Francesco
  Locatello, Peter Gehler, Thomas Brox, Matthias Kümmerer, Matthias Bethge,
  and Bernhard Schölkopf.
\newblock Unsupervised object learning via common fate.
\newblock \emph{arXiv preprint arXiv:2110.06562}, 2021.

\bibitem[Tong et~al.(2022)Tong, Song, Wang, and Wang]{tong2022videomae}
Zhan Tong, Yibing Song, Jue Wang, and Limin Wang.
\newblock Videomae: Masked autoencoders are data-efficient learners for
  self-supervised video pre-training.
\newblock \emph{Advances in Neural Information Processing Systems},
  35:\penalty0 10078--10093, 2022.

\bibitem[Tran et~al.(2022)Tran, Vo, Yamazaki, Fernandes, Kidd, and
  Le]{tran2022aisformer}
Minh Tran, Khoa Vo, Kashu Yamazaki, Arthur Fernandes, Michael Kidd, and Ngan
  Le.
\newblock Aisformer: Amodal instance segmentation with transformer.
\newblock \emph{arXiv preprint arXiv:2210.06323}, 2022.

\bibitem[Tran et~al.(2024)Tran, Vo, Ho, Nguyen, and Le]{tran2024amodal}
Minh Tran, Khoa Vo, Vuong Ho, Tri Nguyen, and Ngan~Hoang Le.
\newblock Amodal instance segmentation with diffusion shape prior estimation.
\newblock In \emph{The First Workshop on Populating Empty Cities--Virtual
  Humans for Robotics and Autonomous Driving at CVPR 2024}, 2024.

\bibitem[Xiao et~al.(2021)Xiao, Xu, Zhong, Luo, Li, and Gao]{xiao2021vrsp}
Yuting Xiao, Yanyu Xu, Ziming Zhong, Weixin Luo, Jiawei Li, and Shenghua Gao.
\newblock Amodal segmentation based on visible region segmentation and shape
  prior.
\newblock In \emph{Proceedings of the AAAI Conference on Artificial
  Intelligence}, pages 2995--3003, 2021.

\bibitem[Xie et~al.(2024)Xie, Yao, Voleti, Jiang, and Jampani]{xie2024sv4d}
Yiming Xie, Chun-Han Yao, Vikram Voleti, Huaizu Jiang, and Varun Jampani.
\newblock Sv4d: Dynamic 3d content generation with multi-frame and multi-view
  consistency.
\newblock \emph{arXiv preprint arXiv:2407.17470}, 2024.

\bibitem[Xu et~al.(2024)Xu, Zhang, and Shi]{xu2024amodal}
Katherine Xu, Lingzhi Zhang, and Jianbo Shi.
\newblock Amodal completion via progressive mixed context diffusion.
\newblock In \emph{Proceedings of the IEEE/CVF Conference on Computer Vision
  and Pattern Recognition}, pages 9099--9109, 2024.

\bibitem[Yang et~al.(2024)Yang, Kang, Huang, Zhao, Xu, Feng, and
  Zhao]{yang2024depthanythingv2}
Lihe Yang, Bingyi Kang, Zilong Huang, Zhen Zhao, Xiaogang Xu, Jiashi Feng, and
  Hengshuang Zhao.
\newblock Depth anything v2.
\newblock \emph{arXiv preprint arXiv:2406.09414}, 2024.

\bibitem[Yao et~al.(2022)Yao, Hong, Wang, Xiao, He, Locatello, Wipf, Fu, and
  Zhang]{yao2022savos}
Jian Yao, Yuxin Hong, Chiyu Wang, Tianjun Xiao, Tong He, Francesco Locatello,
  David~P. Wipf, Yanwei Fu, and Zheng Zhang.
\newblock Self-supervised amodal video object segmentation.
\newblock \emph{Advances in Neural Information Processing Systems},
  35:\penalty0 6278--6291, 2022.

\bibitem[Zhan et~al.(2023)Zhan, Zheng, Xie, and Zisserman]{zhan2023general}
Guanqi Zhan, Chuanxia Zheng, Weidi Xie, and Andrew Zisserman.
\newblock A general protocol to probe large vision models for 3d physical
  understanding.
\newblock \emph{arXiv preprint arXiv:2310.06836}, 2023.

\bibitem[Zhan et~al.(2024)Zhan, Zheng, Xie, and Zisserman]{zhan2024sdamodal}
Guanqi Zhan, Chuanxia Zheng, Weidi Xie, and Andrew Zisserman.
\newblock Amodal ground truth and completion in the wild.
\newblock In \emph{Proceedings of the IEEE/CVF Conference on Computer Vision
  and Pattern Recognition}, pages 28003--28013, 2024.

\bibitem[Zhan et~al.(2020)Zhan, Pan, Dai, Liu, Lin, and Loy]{zhan2020pcnet}
Xiaohang Zhan, Xingang Pan, Bo Dai, Ziwei Liu, Dahua Lin, and Chen~Change Loy.
\newblock Self-supervised scene de-occlusion.
\newblock In \emph{Proceedings of the IEEE/CVF Conference on Computer Vision
  and Pattern Recognition}, pages 3784--3792, 2020.

\bibitem[Zhang et~al.(2023)Zhang, Rao, and Agrawala]{zhang2023controlnet}
Lvmin Zhang, Anyi Rao, and Maneesh Agrawala.
\newblock Adding conditional control to text-to-image diffusion models.
\newblock In \emph{Proceedings of the IEEE/CVF International Conference on
  Computer Vision}, pages 3836--3847, 2023.

\bibitem[Zhu et~al.(2017)Zhu, Tian, Metaxas, and Dollár]{zhu2017semantic}
Yan Zhu, Yuandong Tian, Dimitris Metaxas, and Piotr Dollár.
\newblock Semantic amodal segmentation.
\newblock In \emph{Proceedings of the IEEE Conference on Computer Vision and
  Pattern Recognition}, pages 1464--1472, 2017.

\end{thebibliography}
}

\clearpage
\appendix

\begin{center}
    {\Large\bf Appendix}
\end{center}

In this appendix, we extend the discussion of our approach on video amodal segmentation. We first discuss additional setup details for our method (Sec. \ref{sec:supp-setup}), and then cover more experimental analysis (Sec. \ref{sec:supp-expts}), followed by examples of our method's potential applications (Sec. \ref{sec:supp-applications}). We also show more qualitative results from our method (Sec. 
 \ref{sec:supp-qual}). Please see the project page for a video version of all figures.

\section{Additional setup details}
\label{sec:supp-setup}

\subsection{Inference details}
During inference with our video diffusion model, we follow common practices~\cite{svd} by employing the stochastic sampler from EDM~\cite{edm}. We simplify this process by omitting the second-order correction and keeping the explicit Langevin-like ``churn" factors constant. The denoising process is performed over 25 steps. Specifically, when denoising the latents from $z_t$ to $z_0$ for $i \in \{ t, \cdots, 1 \}$, each denoising step can be expressed as:
\begin{equation}
\mathbf{\hat{z}}_{i-1} \leftarrow \mathbf{\hat{z}}_{i} + (\sigma_{i-1} - \sigma_{i}) \frac{(\mathbf{\hat{z}}_{i}-D_{\theta}(\hat{\mathbf{z}}_{i};\sigma_i))}{\sigma_i}
\label{eq3}
\end{equation}

Furthermore, we employ classifier-free guidance (CFG)~\cite{cfg} to balance the quality and diversity of the generated samples. During training, we randomly set the conditioning to zero with a probability of $\rho = 0.1$ to simulate the unconditional case. During inference, we combine the conditional and unconditional predictions using a guidance scale of $s=1.5$, as defined as:
\begin{equation}
\Tilde{F}_{\theta}(\mathbf{z}, \mathbf{c})=F_{\theta}(\mathbf{z}, \emptyset)\\
+s(F_{\theta}(\mathbf{z}, \mathbf{c})-F_{\theta}(\mathbf{z}, \emptyset))
\label{eq4}
\end{equation}

After denoising, the latent predictions are projected back into pixel space using the VAE decoder, which yields three-channel representations. To convert these into single-channel binary masks in the amodal segmentation stage, we sum the channel values (from 0 to 255) and binarize the predictions by thresholding. The threshold is chosen as a per channel pixel-value of 200. Finally, we take the union of the prediction with the input modal masks, ensuring modal masks remain a subset of amodal masks and are properly reflected in the output.

\subsection{Baselines}
In this section, we provide additional details of the image- and video-level amodal segmentation methods used for comparison.

For image-level amodal segmentation, ‘Convex’~\cite{zhan2020pcnet} generates the geometric convex hull of modal masks, while ‘Convex$^R$’~\cite{zhan2020pcnet} refines this by including only the convex hull within occluded regions predicted by ‘PCNet-M’. ‘PCNet-M’~\cite{zhan2020pcnet} is a self-supervised regression method that recovers amodal masks within occluder areas based on frame-level object ordering recovery. 'AISFormer'~\cite{tran2022aisformer} employs a transformer-based head appended to a modal segmentation backbone to directly predict all amodal bounding boxes and masks within an image. ‘pix2gestalt,’~\cite{ozguroglu2024pix2gestalt} is an image diffusion-based method that generates amodal content conditioned on the RGB image and modal masks of the objects.

For video-level amodal segmentation, ‘SaVos’~\cite{yao2022savos} employs a CNN-LSTM architecture that processes RGB and modal mask patches, along with optical flow, to predict amodal masks and motions. ‘EoRaS’~\cite{fan2023eoras} proposes an object-attention encoder that incorporates Bird’s-Eye View (BEV) 3D information, relying on having access to groundtruth camera parameters. ‘C2F-Seg’~\cite{gao2023c2fseg} leverages a vector-quantized latent space for coarse feature learning, refined with a convolutional module; though designed for image-level tasks, it extends to video segmentation using a spatial-temporal transformer block.

For generic video regression approaches, ‘VideoMAE’~\cite{tong2022videomae} is a transformer-based autoencoder that we adapt for our task by setting the masking ratio to zero, applying supervised training, and using the decoder during inference. ‘3D-UNet’~\cite{vldm}, the backbone of our video diffusion model, contains interleaved residual and transformer blocks with spatial and temporal modules but is trained to perform one-step generation without any iterative denoising.

\section{Additional experiments}
\label{sec:supp-expts}

Note that the video versions of all qualitative results in this and the following sections can be found directly on the project page.

\paragraph{Improved results on MOVi-B/D.}
All results we report till now on MOVi-B/D follow prior work in segmenting objects in a region which is defined as a 100\% extension of the region enclosed by the input modal mask. Therefore, all images are cropped to this region before being sent as input to any of the methods. This is different from the standard protocol used in other datasets, where the \textit{entire} image is sent as input to the methods (without any cropping). Here, we include results from training our model with the entire image as input on the MOVi-B/D datasets. 
As shown in Table~\ref{tab:uncropped_movi}, this fix significantly improves metrics, with our method achieving 4\% and 6\% gains in mIoU on MOVi-B and MOVi-D, respectively. Regression methods also benefit notably from this setting.
We conclude that this is because MOVi-B/D include many instances of complete occlusions of objects, for which segmentation in a cropped region is not enough for predicting the amodal mask.

\begin{table}[t]
  \centering
  \caption{\textbf{Quantitative results on MOVi-B/D with uncropped input.} Enlarged modal region-cropped input limits the model’s ability to predict an amodal mask when an object is fully occluded. Using the entire image as input restores the model's ability to complete amodal masks fully, especially when the modal area is small. This results in substantial metric improvements compared to Table 2 in the main paper. We copy over the results here for reference.} 
  \scriptsize
  \begin{tabular}{@{}lccccccc@{}}
    \toprule
     \multirow{2}{*}{Input} & \multirow{2}{*}{Method}  & \multicolumn{2}{c}{MOVi-B} &  \multicolumn{2}{c}{MOVi-D} \\ & 
    & mIoU & mIoU$_{occ}$ & mIoU & mIoU$_{occ}$ \\
    \midrule

    \multirow{4}{*}{\shortstack{modal\\cropped}} & VideoMAE~\cite{tong2022videomae} & 78.74 & 42.86 & 70.93 & 32.78 \\
    & 3D-UNet & 82.16 & 49.81 & 75.65 & 40.86 \\
   & Ours (Top-1) & 83.51 & 53.75 & 77.03 & 44.23 \\
   & Ours (Top-3) & 83.93 & 54.56 & 77.76 & 45.6 \\
    
    \midrule
    
    \multirow{4}{*}{uncropped} & VideoMAE~\cite{tong2022videomae}  & 85.35 & 49.53 & 79.13 & 42.41 \\
    & 3D-UNet &  84.24 & 46.17 & 76.90 & 36.69 \\
    & Ours (Top-1) & \underline{87.8} & \underline{53.69} & \underline{82.97} & \underline{47.86} \\
    & Ours (Top-3) & \textbf{88.43} & \textbf{54.64} & \textbf{84.04} & \textbf{49.43} \\
    \bottomrule
  \end{tabular}
  \label{tab:uncropped_movi}
\end{table}

\begin{figure}[t]
  \centering
   \includegraphics[width=1.0\linewidth]{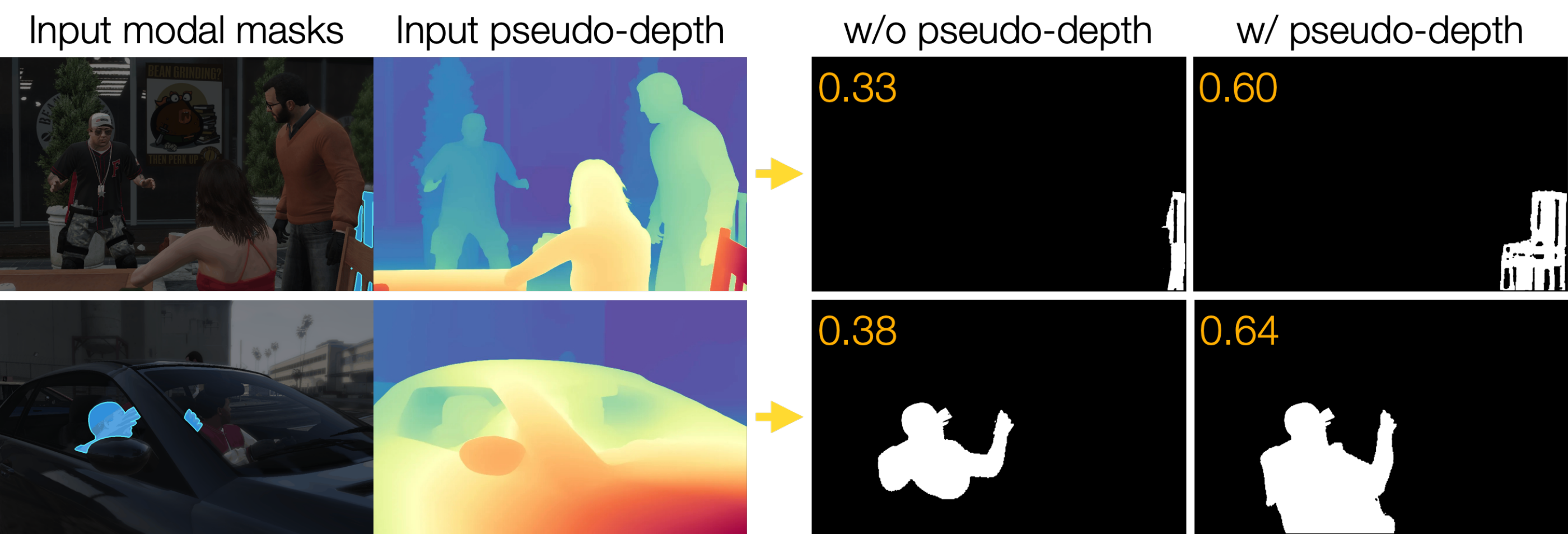}

   \caption{We show \textbf{how pseudo-depth aids amodal segmentation}. Object's surrounding regions with lower depth values, i.e., closer to the camera, act as potential occluders. In the top row, the occluders are the person and chair to the left of the object; in the bottom row, the occluder is the car door below the person. Depth information implicitly guides our method to complete these occluded regions.}
   \label{fig:depth_cond}
\end{figure}

\paragraph{Qualitative evidence for pseudo-depth conditioning.}
The quantitative advantage of pseudo-depth conditioning was demonstrated in Table 3 of the main paper. Here, we provide qualitative evidence to illustrate the source of this improvement. As shown in Figure~\ref{fig:depth_cond}, pseudo-depth conditioning encourages our method to segment areas \textit{closer} to the camera, suggesting that depth serves as an implicit indicator of potential occluders and therefore, gives information about which occluded boundary to extend in order to predict the amodal mask.

\paragraph{Ablation on weights initialization.} 
We leverage the real-world priors learnt by large-scale diffusion models by utilizing pretrained SVD checkpoints~\cite{svd}. Here, we evaluate the importance of this initialization. In Table~\ref{tab:ablation_svd_prior}, we compare the performance of our model and the 3D U-Net with and without pretrained weights. Results show that excluding the checkpoint leads to a performance drop for both models, with a more pronounced decline for ours. These results underscore the importance of the SVD priors.

\paragraph{Building an end-to-end segmentation and completion model} 
Unlike our two-stage method, which first performs amodal segmentation and then inpaints content, the image diffusion-based method pix2gestalt~\cite{ozguroglu2024pix2gestalt} adopts a one-stage approach to directly generate amodal content and derive masks. A similar one-stage approach can be extended for our video setting. However, as shown in Table~\ref{tab:ablation_one_stage}, our two-stage method demonstrates clear advantages over the one-stage approach. We attribute this low performance of the end-to-end method to the lack of data available for training such a single-stage method.
In contrast, the two-stage method benefits from breaking down the pipeline into video amodal segmentation and content completion. For the former, it is easy to find large-scale training data of modal-amodal mask pairs from synthetic datasets. For the latter, since the content completion task reduces to video inpainting, less amount of training data is sufficient for finetuning.

\begin{table}[t]
  \centering
  \caption{\textbf{Ablation of SVD priors.} We study the effect of using pretrained SVD weights as initialization for our training. We find that leveraging priors from large-scale pretraining of SVD enhances both our method and the 3D UNet baseline, with particularly substantial improvements observed for our method. }
  \footnotesize
  \begin{tabular}{@{}cccccccc@{}}
    \toprule
    \multirow{2}{*}{Method} & \multirow{2}{*}{\makecell{pretrained \\ ckpt?}} & \multicolumn{2}{c}{SAIL-VOS} & \multicolumn{3}{c}{TAO-Amodal}\\
    &  & mIoU & mIoU$_{occ}$ & AP$_{25}$ & AP$_{50}$ & AP$_{75}$ \\
    \midrule
    Ours  & \ding{55} & 68.89 & 26.96 & 93.73 & 79.45 & 57.87 \\
    Ours &  \ding{51} & \textbf{75.17} & \textbf{51.28} & \textbf{94.89} & \textbf{85.03} & \textbf{66.87} \\
    \midrule
    3D UNet & \ding{55} & 70.85 & 32.66 & \textbf{94.88} & 83.81 & 59.75 \\
    3D UNet & \ding{51} & \textbf{72.79} & \textbf{39.54} & 94.59 & \textbf{83.83} & \textbf{64.33} \\
    \bottomrule
  \end{tabular}
  \label{tab:ablation_svd_prior}
\end{table}

\begin{table}[t]
  \centering
  \caption{\textbf{Ablation study on end-to-end amodal content completion.} We train an end-to-end version of our two-stage pipeline with a dataset of curated modal-amodal RGB training pairs from SAIL-VOS, in a similar fashion to pix2gestalt \cite{ozguroglu2024pix2gestalt}. Compared to the two-stage results in Table 1 of the main paper, this approach shows a significant performance drop in both in-domain and zero-shot evaluations, highlighting the superiority of the two-stage method.}
  \small
  \begin{tabular}{@{}ccccccc@{}}
    \toprule
    \multirow{2}{*}{Method} & \multicolumn{2}{c}{SAIL-VOS} & \multicolumn{3}{c}{TAO-Amodal}\\
    & mIoU & mIoU$_{occ}$ & AP$_{25}$ & AP$_{50}$ & AP$_{75}$ \\
    \midrule
    Two-stage & \textbf{77.07} & \textbf{55.12} & \textbf{97.28} & \textbf{89.25} & \textbf{71.99} \\
    One-stage  & 66.15 & 40.31 & 70.65 & 57.51 & 37.22 \\
    \bottomrule
  \end{tabular}
  \label{tab:ablation_one_stage}
\end{table}

\section{Examples of applications}
\label{sec:supp-applications}
\begin{figure}[t]
  \centering
   \includegraphics[width=1.0\linewidth]{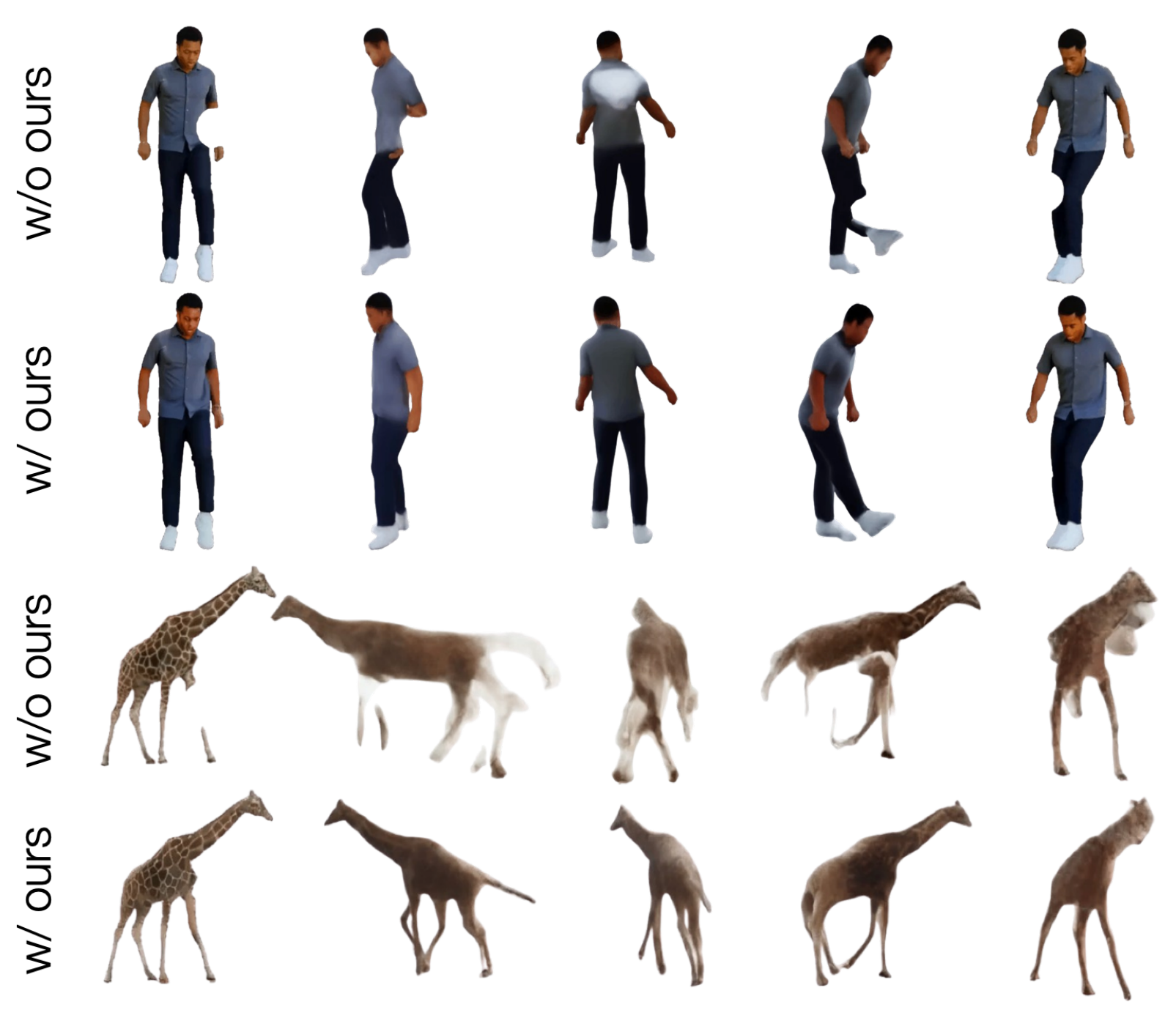}

   \caption{\textbf{4D reconstruction results}.  Without amodal completion by our method, the 4D reconstruction exhibits blank regions and unrealistic artifacts in occluded areas, such as the person’s back and leg. The varying occluded portions over time confuse SV4D, disrupting its understanding of the object's 3D structure. In contrast, using completed objects from our method significantly improves the reconstruction quality, producing more consistent and clear novel-views.}
   \label{fig:4d_recon}
\end{figure}

\paragraph{4D reconstruction.}
Our method enables 4D reconstruction for occluded objects when used in conjunction with off-the-shelf SV4D~\cite{xie2024sv4d}. In Figure~\ref{fig:4d_recon}, we compare reconstructions with and without completion. Without completion, blank regions appear in occluded areas, making it more difficult to hallucinate reasonable re-projections across different views. In contrast, our method allows SV4D to produce consistent and clearer 4D reconstructions.

\begin{figure}[t]
  \centering
   \includegraphics[width=1.0\linewidth]{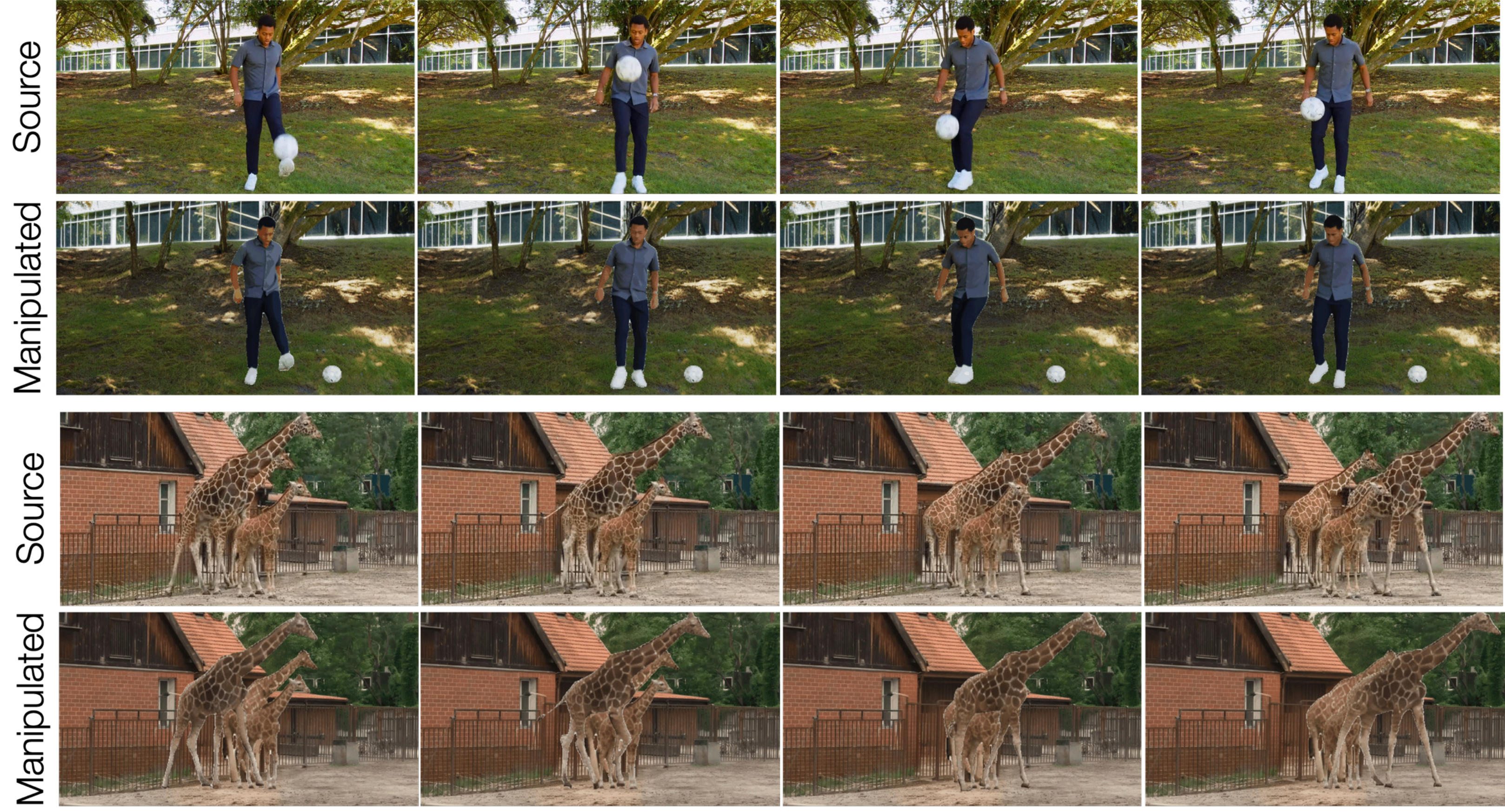}

   \caption{\textbf{Scene manipulation examples}. Using de-occluded objects from our method, we can reposition and reorder them to create new scenes. In the top rows, the relationship between the person and the soccer ball is altered, changing the scene from “the person is juggling” to “the person places the soccer ball aside and practices a juggling posture.” In the bottom rows, the middle giraffe is moved to the front and its position is adjusted. }
   \label{fig:scene_manip}
\end{figure}

\paragraph{Scene manipulation.}
With amodally completed objects in the scene, we can change their orderings and positions without exposing previously occluded regions. Figure~\ref{fig:scene_manip} shows examples of scene manipulation, where our method facilitates manual re-composition of scenes by inpainting the occluded content of objects.

\begin{figure*}[h]
  \centering
   \includegraphics[width=0.87\linewidth]{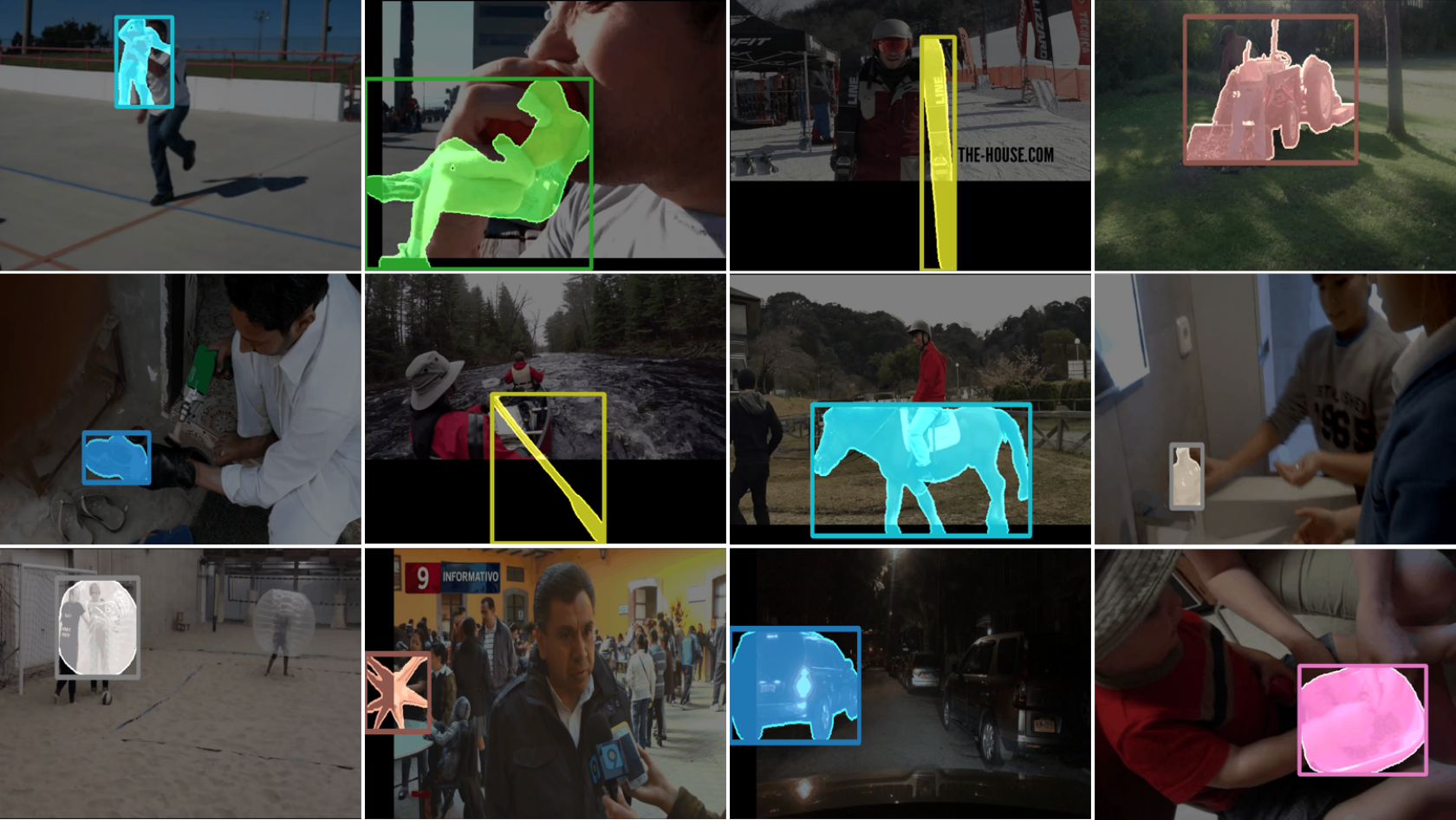}

   \caption{\textbf{Qualitative results for pseudo-ground truth of TAO-Amodal masks}.  Leveraging the amodal bounding box as a strong prior, our method demonstrates versatility across diverse categories, such as person, tractor, and bottles, and generalizes well to unseen categories like snowboards and horses. This high-quality pseudo-ground truth can semi-automate the manual annotation of amodal masks in real-world videos.}
   \label{fig:pseudo_gt}
\end{figure*}

\paragraph{Pseudo-groudtruth for TAO-Amodal masks.} 
TAO-Amodal~\cite{hsieh2023taoamodal} provides ground truth for amodal bounding boxes but lacks annotations for amodal \textit{masks} due to the challenges of manual labeling of occluded objects in videos. We show that our method can be used to generate high-quality \textit{pseudo}-ground truth masks for this dataset by using the information about ground-truth amodal bounding boxes, which define the extent of the amodal shape. We find that using the amodal bounding boxes to crop the input modal mask sequences, one can train a more accurate video amodal segmentation method exclusively on SAIL-VOS. This way, our approach significantly improves evaluation metrics and aligns precisely with the amodal bounding box extent, as shown in Table~\ref{tab:pseudo_gt}. Figure~\ref{fig:pseudo_gt} further illustrates the qualitative results of the pseudo-ground truth masks which are high-fidelity across diverse object categories. Quantitatively, we find that using the pseudo-groundtruths for finetuning baselines like VideoMAE (which have already been pre-trained on SAIL-VOS), improves their performance on the TAO-Amodal dataset by around 25\%, 25\%, and 20\% on AP$_{25}$, AP$_{50}$, and AP$_{75}$ respectively. Apart from this, the generated pseudo-groundtruths can be used to semi-automate the amodal mask annotation process as this is a challenging and inherently ill-posed problem.

Note that we do not include this data point in the main paper as at inference we cannot expect to have access to amodal bounding boxes but in order to produce pseudo-groundtruth, one can adopt this approach.

\begin{table}[t]
  \centering
  \caption{\textbf{Pseudo-groundtruths on TAO-Amodal.} We show that using the amodal bounding box prior from the TAO-Amodal dataset to specify the extent of the output amodal segmentation mask, can help improve the quality of video amodal segmentation. We use this version of our method to produce `pseudo-groundtruths' for TAO-Amodal. We find that these pseudo-annotations can help improve the quantitative performance of baselines like VideoMAE. See text for more details}
  \small
  \begin{tabular}{@{}ccccccc@{}}
    \toprule
    \multirow{2}{*}{Input setting} & \multicolumn{2}{c}{SAIL-VOS} & \multicolumn{3}{c}{TAO-Amodal}\\
    & mIoU & mIoU$_{occ}$ & AP$_{25}$ & AP$_{50}$ & AP$_{75}$ \\
    \midrule
    Uncropped & 77.07 & 55.12 & 97.28 & 89.25 & 71.99 \\
    Amodal cropped  & \textbf{87.44} & \textbf{69.81} & \textbf{99.59} & \textbf{99.59} & \textbf{99.48} \\
    \bottomrule
  \end{tabular}
  \label{tab:pseudo_gt}
\end{table}

\begin{figure*}[t]
  \centering
   \includegraphics[width=\linewidth]{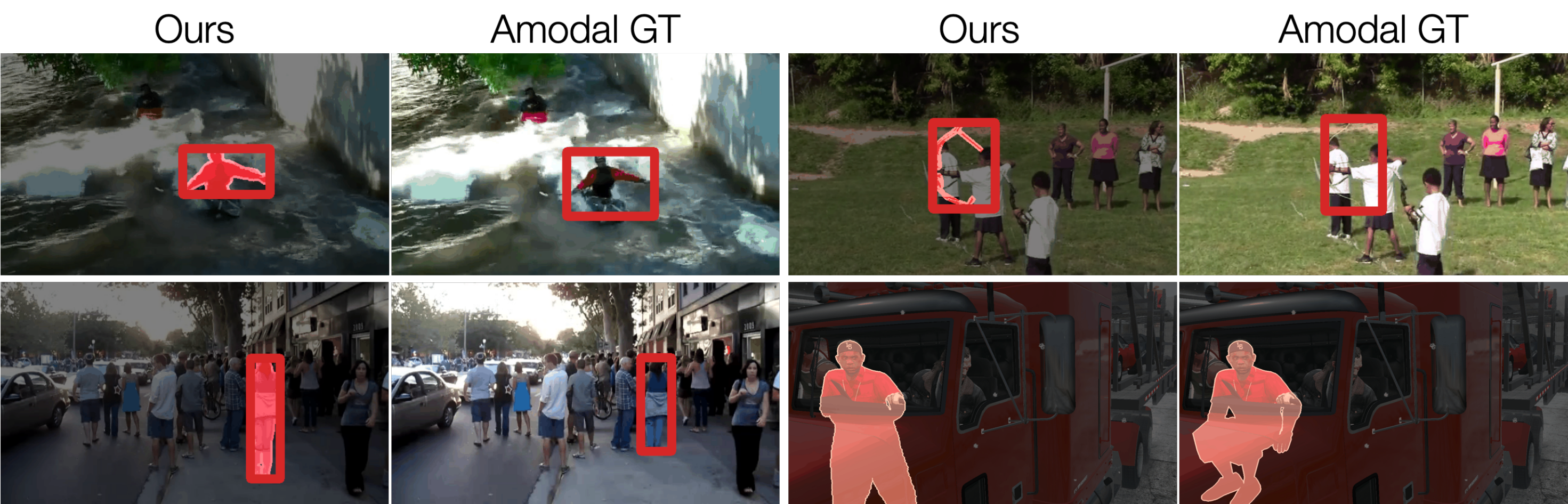}
   \caption{Qualitative analysis of failure cases of our method. See text for more details.}
   \label{fig:supp_limitation}
\end{figure*}

\section{Qualitative results}
\label{sec:supp-qual}

A video version of all figures in this section are available on the project page.

\begin{figure*}[t]
  \centering
   \includegraphics[width=0.9\linewidth]{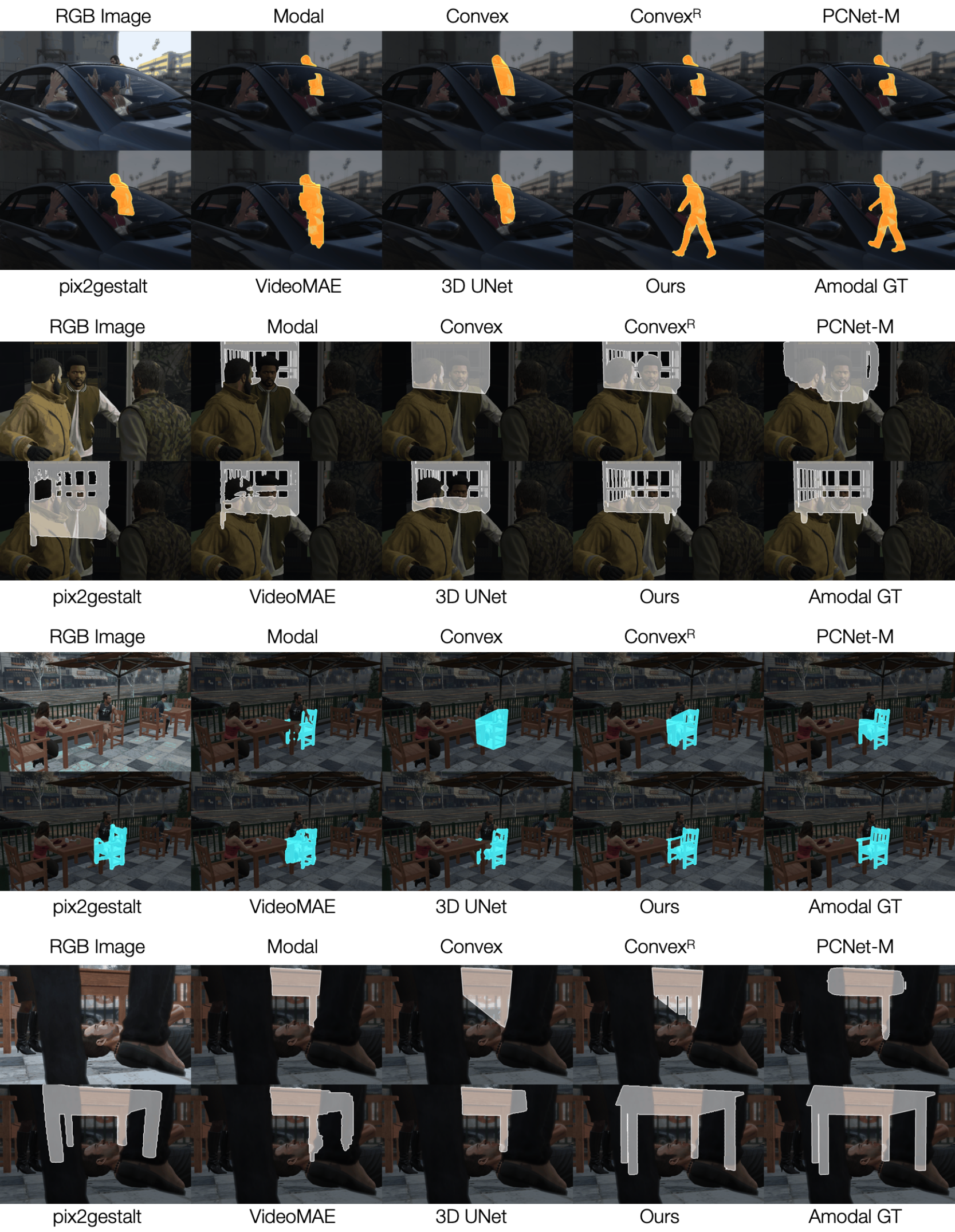}
   \caption{Qualitative results on SAIL-VOS. (1/2)}
   \label{fig:supp_sailvos1}
\end{figure*}

\begin{figure*}[t]
  \centering
   \includegraphics[width=0.85\linewidth]{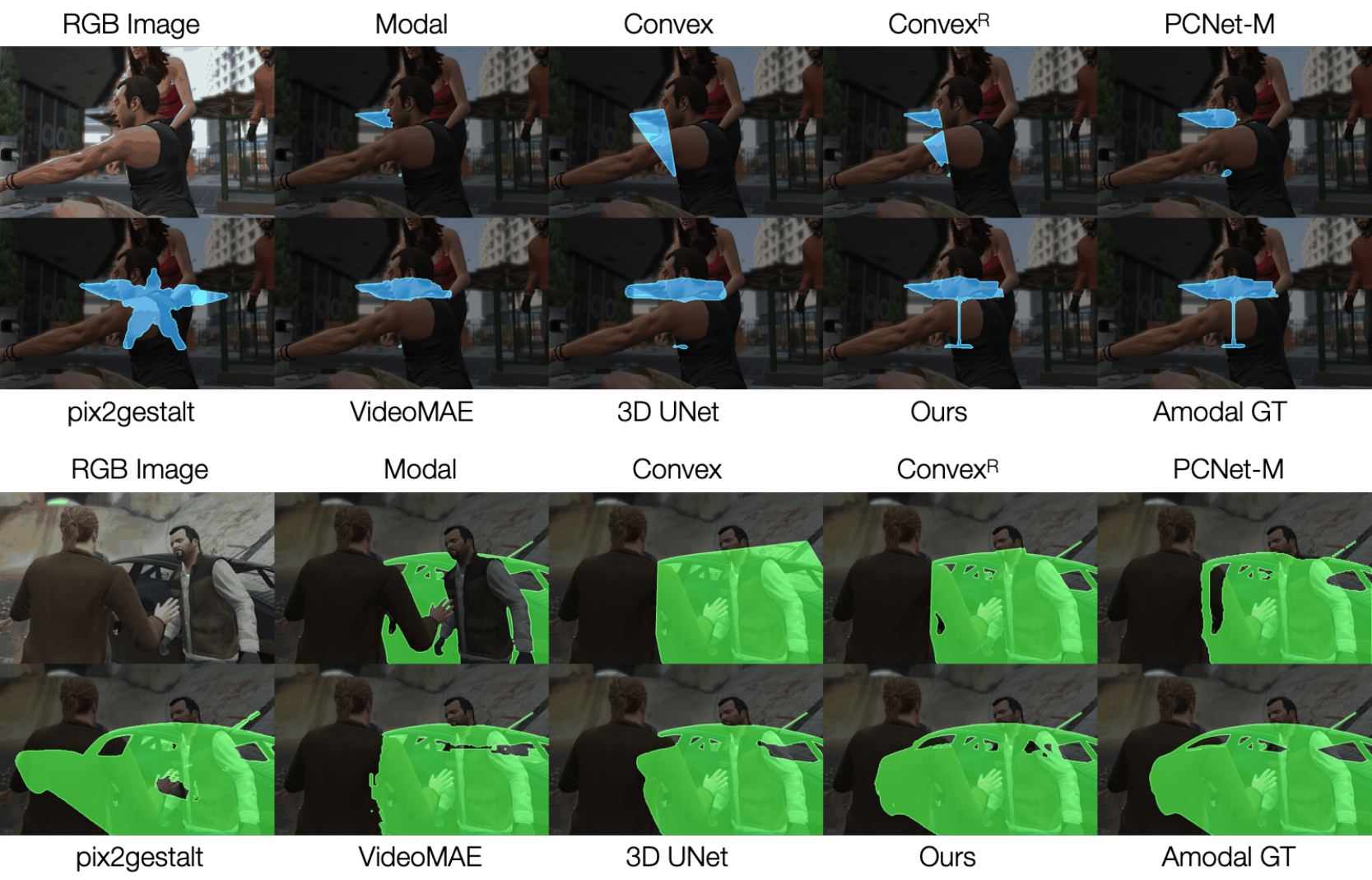}
   \caption{Qualitative results on SAIL-VOS. (2/2)}
   \label{fig:supp_sailvos2}
\end{figure*}

\begin{figure*}[t]
  \centering
   \includegraphics[width=0.85\linewidth]{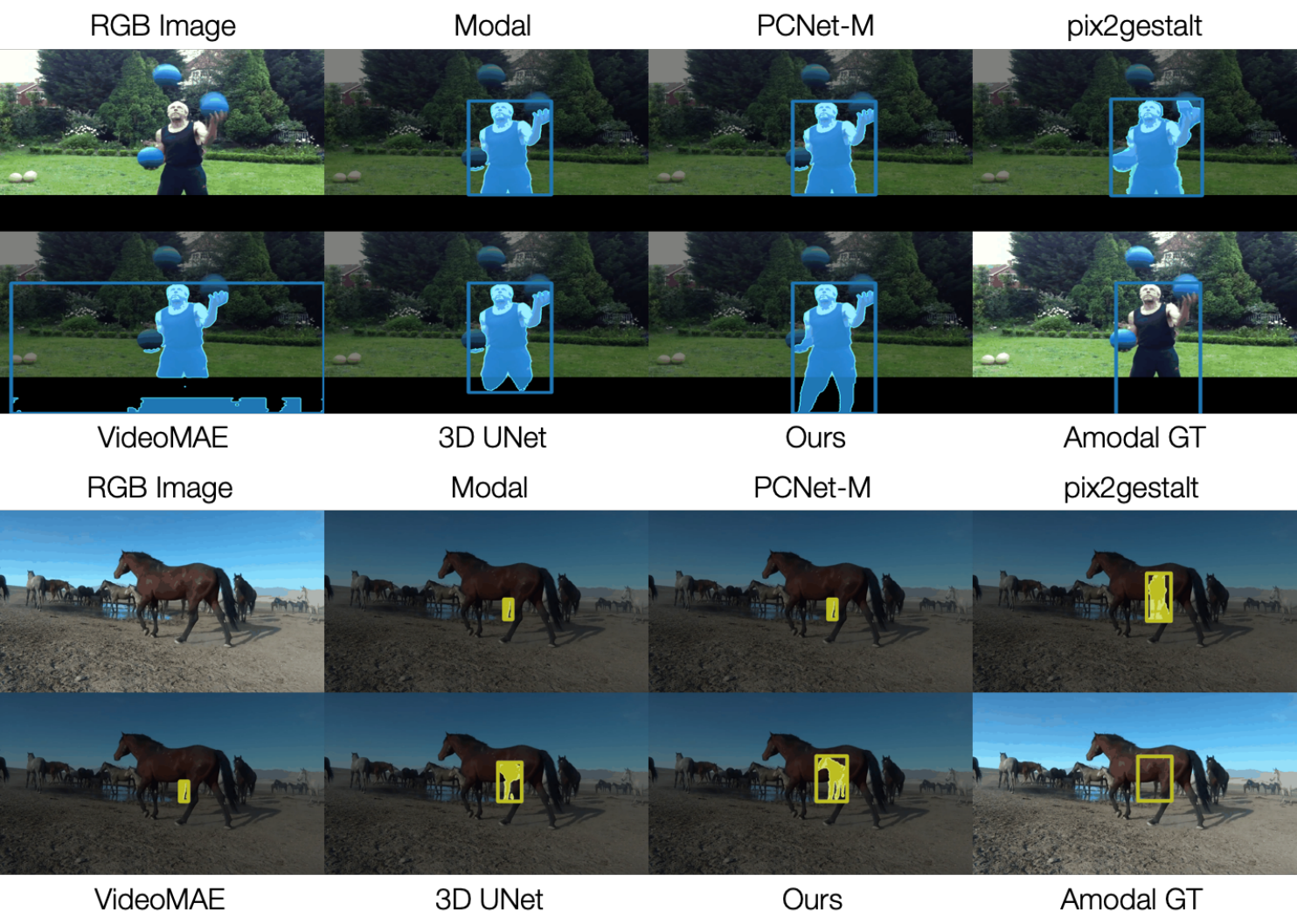}
   \caption{Qualitative results on TAO-Amodal. (1/2)}
   \label{fig:supp_tao1}
\end{figure*}

\begin{figure*}[t]
  \centering
   \includegraphics[width=0.9\linewidth]{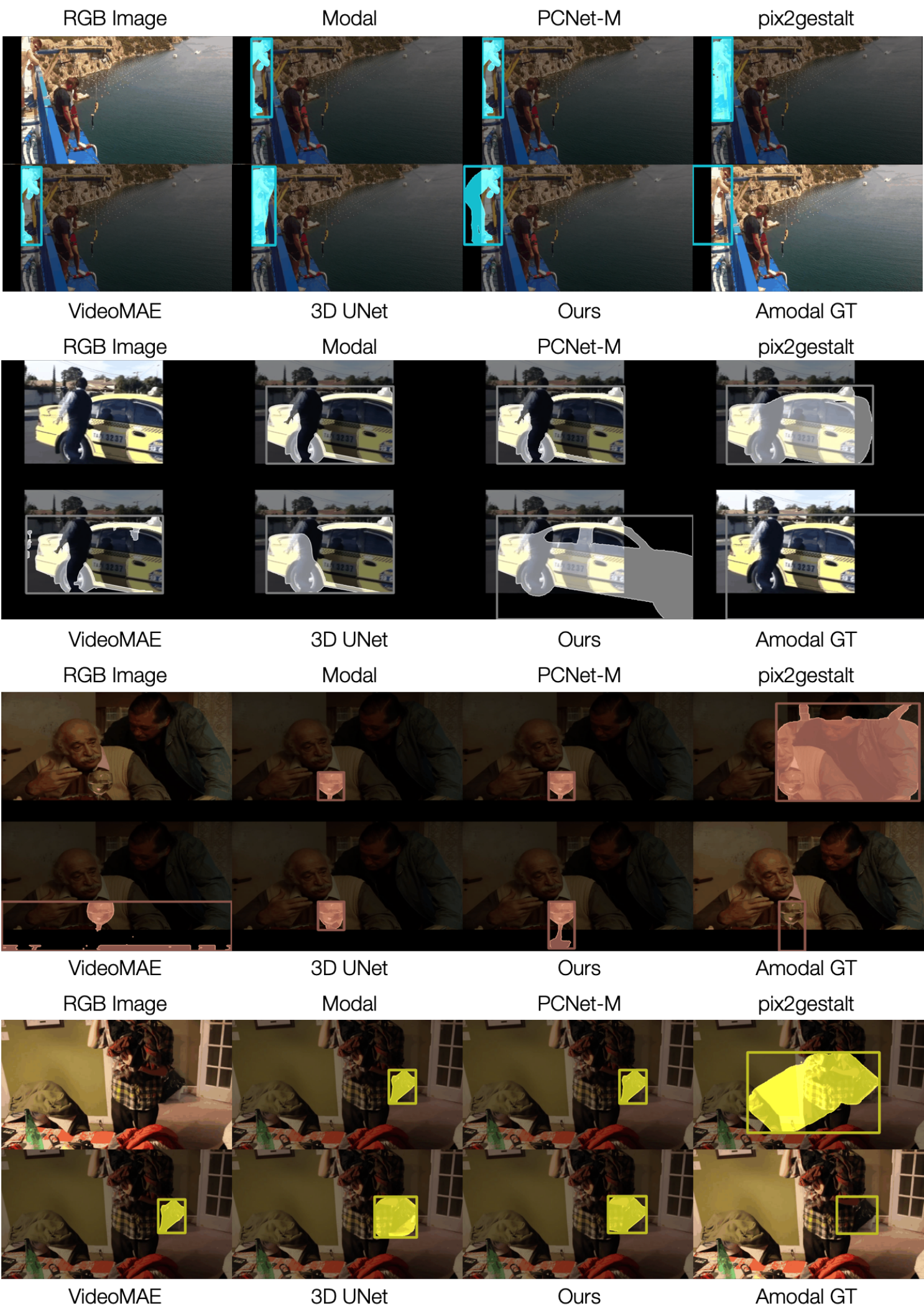}
   \caption{Qualitative results on TAO-Amodal. (2/2)}
   \label{fig:supp_tao2}
\end{figure*}

\begin{figure*}[t]
  \centering
   \includegraphics[width=1\linewidth]{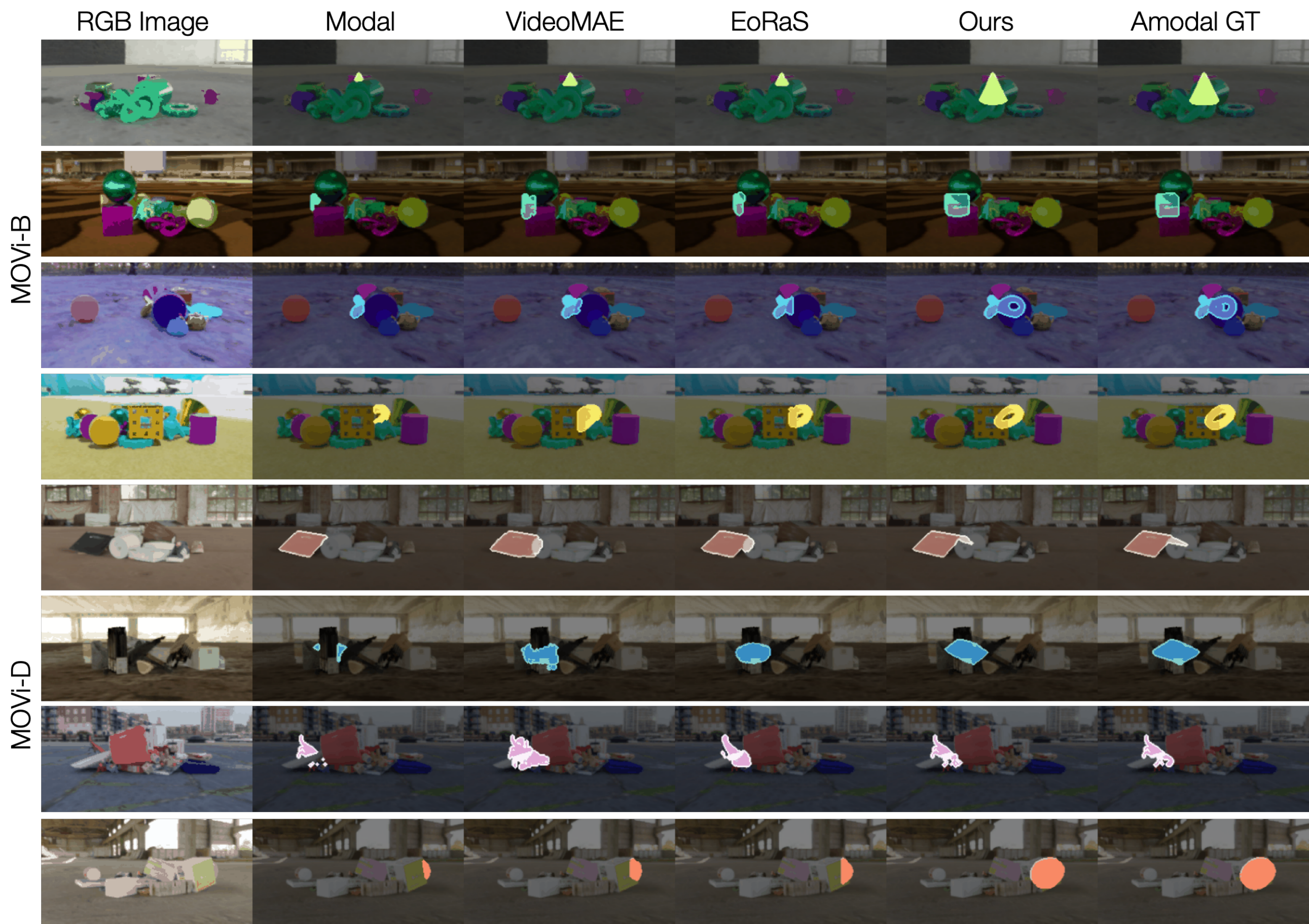}
   \caption{Qualitative results on MOVi-B/D.}
   \label{fig:supp_movi}
\end{figure*}

\begin{figure*}[t]
  \centering
   \includegraphics[width=1\linewidth]{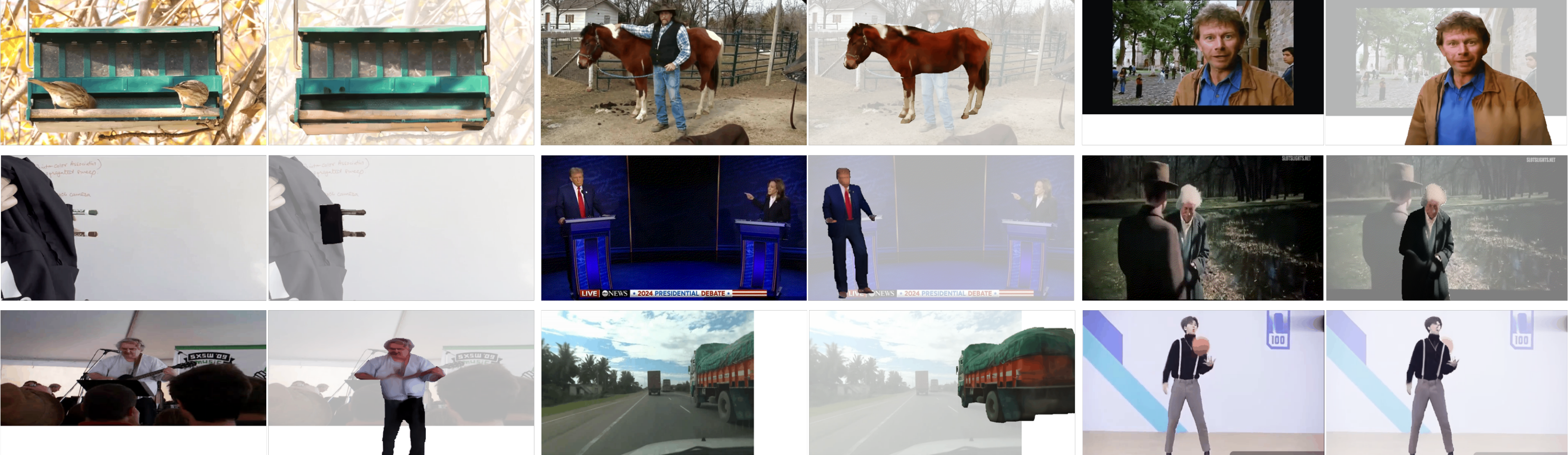}
   \caption{Qualitative results for amodal content completion for in-the-wild scenarios.}
   \label{fig:supp_content_comp}
\end{figure*}

Here, we present qualitative results from all datasets and additional, in-the-wild scenarios. Figures \ref{fig:supp_sailvos1}, \ref{fig:supp_sailvos2}, \ref{fig:supp_tao1}, \ref{fig:supp_tao2} and \ref{fig:supp_movi} compare our amodal segmentation method with more baselines on SAIL-VOS, TAO-Amodal, and MOVi-B/D. Our method demonstrates superior performance in generating high-fidelity shapes in the occluded regions of objects. Figure \ref{fig:supp_content_comp} showcases additional in-the-wild content completion results, highlighting the photo-realistic quality and strong generalization capability of our method.

\paragraph{Failure cases.} In Figure \ref{fig:supp_limitation}, we show four different kinds of failure cases. In the first case with a person swimming, our method does not successfully complete the person's amodal region. This happens often if the object of interest is occluded throughout the extent of the video; our model is not able to understand if this is a completely visible object or a consistently occluded object. 
In the second case, the occluded object is a bow, which has never been seen before and is completely out-of-distribution from the set of objects in SAIL-VOS. Our method fails in this case.
In the third and fourth case, our method incorrectly assumes the height of a completely visible man to be greater than what it is, and predicts a sitting person to be standing. Therefore, our method lacks contextual cues about what the scene is and how the modal region looks like in the first-stage.

\newpage

\end{document}